\documentclass[10pt,journal,compsoc]{IEEEtran}

\ifCLASSOPTIONcompsoc
  \usepackage[nocompress]{cite}
\else
  \usepackage{cite}
\fi

\usepackage[colorlinks,linkcolor=red]{hyperref}
\usepackage{url}            %
\usepackage{booktabs}       %
\usepackage{amsfonts}       %
\usepackage{nicefrac}       %
\usepackage{microtype}      %
\usepackage{xcolor}         %
\usepackage{graphicx}
\usepackage{wrapfig}
\usepackage{multirow}
\usepackage{graphicx}
\usepackage{caption}
\usepackage{float}
\usepackage{amsmath}
\usepackage{bbding}
\usepackage{color}
\usepackage{cuted}
\usepackage{amssymb}
\usepackage{mathtools}
\usepackage{amsthm}

\usepackage{epsfig}
\usepackage{graphicx}
\usepackage{amsmath}
\usepackage{amssymb}
\usepackage{multirow}
\usepackage{enumerate}
\usepackage{bm}
\usepackage{wrapfig}
\usepackage{epstopdf}
\usepackage{wrapfig}
\DeclareMathOperator*{\argmin}{argmin}

\newcommand{\md}[1]{\textcolor[rgb]{0.0,0.0,0.0}{#1}}
\newcommand{\js}[1]{\textcolor[rgb]{0.0,0.0,0.0}{#1}}

\ifCLASSINFOpdf
\else
\fi

\begin{document}

\title{Fast Learning of Signed Distance Functions from Noisy Point Clouds via Noise to Noise Mapping}

\author{Junsheng Zhou\IEEEauthorrefmark{1},
        Baorui Ma\IEEEauthorrefmark{1}, 
        Yu-Shen Liu,~\IEEEmembership{Member,~IEEE,} Zhizhong Han
\IEEEcompsocitemizethanks{
\IEEEcompsocthanksitem Junsheng Zhou, Baorui Ma are with the School of Software, Tsinghua University, Beijing, China. Baorui Ma is also with BAAI, Beijing, China. E-mail: zhoujs21@mails.tsinghua.edu.cn, brma@baai.ac.cn.
\IEEEcompsocthanksitem Yu-Shen Liu is with the School of Software, Tsinghua University, Beijing, China. E-mail: liuyushen@tsinghua.edu.cn
\IEEEcompsocthanksitem Zhizhong Han is with the Department of Computer Science, Wayne State University, USA. E-mail: h312h@wayne.edu
}
\thanks{Junsheng Zhou and Baorui Ma contribute equally to this work. The corresponding author is Yu-Shen Liu. This work was supported by National Key R\&D Program of China (2022YFC3800600), and the National Natural Science Foundation of China (62272263, 62072268). Project page: \url{https://mabaorui.github.io/Noise2NoiseMapping/}.}
}

\IEEEtitleabstractindextext{%
\begin{abstract}
Learning signed distance functions (SDFs) from point clouds is an important task in 3D computer vision. However, without ground truth signed distances, point normals or clean point clouds, current methods still struggle from learning SDFs from noisy point clouds. To overcome this challenge, we propose to learn SDFs via a noise to noise mapping, which does not require any clean point cloud or ground truth supervision. Our novelty lies in the noise to noise mapping which can infer a highly accurate SDF of a single object or scene from its multiple or even single noisy observations. We achieve this by a novel loss which enables statistical reasoning on point clouds and maintains geometric consistency although point clouds are irregular, unordered and have no point correspondence among noisy observations. To accelerate training, we use multi-resolution hash encodings implemented in CUDA in our framework, which reduces our training time by a factor of ten, achieving convergence within one minute. We further introduce a novel schema to improve multi-view reconstruction by estimating SDFs as a prior. Our evaluations under widely-used benchmarks demonstrate our superiority over the state-of-the-art methods in surface reconstruction from point clouds or multi-view images, point cloud denoising and upsampling.
\end{abstract}

}

\maketitle

\IEEEdisplaynontitleabstractindextext

\IEEEpeerreviewmaketitle

\begin{strip}\centering
\vspace{-1.6in}
\includegraphics[width=\textwidth]{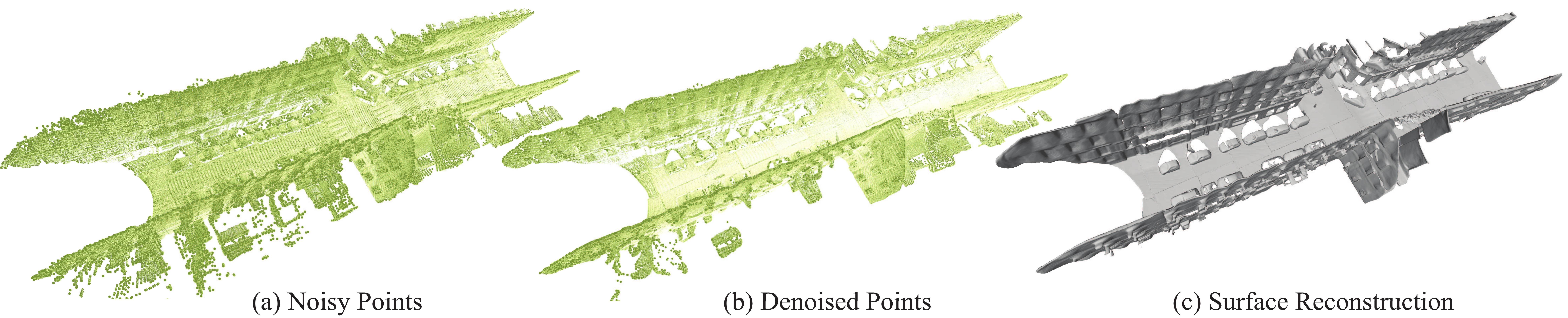}
\captionof{figure}{We introduce to learn signed distance functions (SDFs) for single noisy point clouds. Our method does not require ground truth signed distances, point normals or clean points as supervision for training. We achieve this via learning a mapping from one noisy observation to another or even on a single observation. Our novel learning manner is supported by modern Lidar systems which capture 10 to 30 noisy observations per second. We show the SDF learned from (a) a single real scan containing $10M$ points, (b) the denoised point cloud and (c) the reconstructed surface. Fig.~\ref{fig:Paris} demonstrates our superiority over the latest surface reconstructions in this case.
\label{fig:Paris1}}
\end{strip}

\section{Introduction}
3D point clouds have been a popular 3D representation. We can capture 3D point clouds not only on unmanned vehicles, such as self-driving cars, but also from consumer level digital devices in our daily life, such as the iPhone. However, the raw point clouds are discretized and noisy, which is not friendly to downstream applications like virtual reality and augmented reality requiring clean surfaces. This results in a large demand of learning signed distance functions (SDFs) from 3D point clouds, since SDFs are continuous and also capable of representing arbitrary 3D topology.

Deep learning based methods have shown various solutions of learning SDFs from point clouds~\cite{DBLP:conf/icml/GroppYHAL20,Atzmon_2020_CVPR,Zhizhong2021icml,jiang2020lig,Peng2021SAP}. Different from classic methods~\cite{DBLP:journals/tog/KazhdanH13,OhtakeBATS03}, they mainly leverage data-driven strategy to learn various priors from large scale dataset using deep neural networks. They usually require the signed distance ground truth~\cite{Liu2021MLS}, point normals~\cite{jiang2020lig,DBLP:conf/eccv/ChabraLISSLN20,Peng2021SAP}, additional constraints~\cite{DBLP:conf/icml/GroppYHAL20,Atzmon_2020_CVPR} or no noise assumption~\cite{Zhizhong2021icml}. These requirements significantly affect the accuracy of SDFs learned for noisy point clouds, either caused by poor generalization or the incapability of denoising. Therefore, it is still challenging to learn SDFs from noisy point clouds without clean or ground truth supervision.

To overcome this challenge, we introduce to learn SDFs from noisy point clouds via noise to noise mapping. Our method does not require ground truth signed distances, point normals or clean point clouds to learn priors. As demonstrated in Fig.~\ref{fig:Paris1}, our novelty lies in the way of learning a highly accurate SDF for a single object or scene from its several corrupted observations, i.e., noisy point clouds. Our learning manner is supported by modern Lidar systems which produce about 10 to 30 corrupted observations per second. By introducing a novel loss function containing a geometric consistency regularization, we are enabled to learn a SDF via a task of learning a mapping from one corrupted observation to another corrupted observation or even a mapping from one corrupted observation to the observation itself. The key idea of this noise to noise mapping is to leverage the statistical reasoning to reveal the uncorrupted structures upon its several corrupted observations. One of our contribution is the finding that we can still conduct statistical reasoning \js{even when there is} no spatial correspondence among points on different corrupted observations.

\md{We originally presented our method at ICML2023 \cite{BaoruiNoise2NoiseMapping}, and then extend our method in  more applications with novel frameworks. Specifically, 
we get inspiration from Instant-NGP \cite{mueller2022instant} and propose a fast learning framework to leverage multi-resolution hash tables of learnable features. This design improves the efficiency of SDF inference in our noise to noise mapping process, and reduces our training time from 15 minutes to one minute. 
}

We further extend our method in the multi-view setting. 
Recent methods \cite{neuslingjie, yariv2021volume} learn SDFs in multi-view reconstruction by minimizing the volume rendering error introduced in NeRF \cite{mildenhall2020nerf} which learns radiance fields for scene representation. 
Using the SDF estimated by our method as a prior, we improve the multi-view reconstruction accuracy by removing artifacts in empty space.
Specifically, we first leverage Structure from Motion
(SfM) methods (e.g. COLMAP \cite{schoenberger2016sfm}) to estimate a noisy point cloud from multi-view images and learn a signed distance field from it using our fast learning framework. \js{The learned field is then adopted as a prior to guide multi-view reconstruction methods (e.g. NeuS \cite{neuslingjie}) for reconstructing surfaces from multi-view images. The key reason that our method can serve as the guidance for multi-view reconstruction is that our method learns clean and accurate SDFs from noisy point clouds obtained by SfM.}

Our results achieve the state-of-the-art in different applications including surface reconstruction from point clouds and multi-view images, point cloud denoising and upsampling under widely used benchmarks. Our contributions are listed below.

\begin{enumerate}[i)]
\item We introduce a method to learn SDFs from noisy point clouds without requiring ground truth signed distances, point normals or clean point clouds.
\item We prove that we can leverage Earth Mover's Distance (EMD) to perform the statistical reasoning via noise to noise mapping and justify this idea using our novel loss function, even if 3D point clouds are irregular, unordered and have no point correspondence among different observations.
\item \md{We significantly accelerate the training process to enable the convergence of SDFs within one minute by integrating multi-resolution hash encoding into our framework.}
\item \md{We propose a novel schema to use the SDF learned from noisy SfM points as a prior for improving multi-view reconstruction.}
\item \md{We achieved the state-of-the-art results in surface reconstruction from point clouds or multi-view images, point cloud denoising and upsampling for shapes or scenes under widely used benchmarks.}
\end{enumerate}

\section{Related Work}
Learning implicit functions for 3D shapes and scenes has made great progress~\cite{mildenhall2020nerf,Oechsle2021ICCV,handrwr2020,zhizhongiccv2021matching,jin2023multi,zhizhongiccv2021completing,takikawa2021nglod,DBLP:journals/corr/abs-2105-02788,rematasICML21,feng2022np,li2023neaf,li2024LDI,ma2023geodream,chen2023gridpull,koneputugodage2023octree}. We briefly review methods with different supervision below.

\noindent\textbf{Learning from 3D Supervision. }It was explored on how to learn implicit functions, i.e., SDFs or occupancy fields, using 3D supervision including signed distances~\cite{DBLP:journals/corr/abs-1901-06802,Park_2019_CVPR,aminie2022,ma2023towards,huang2022neural,zhu2022semi,zhu2024ssp} and binary occupancy labels~\cite{MeschederNetworks,chen2018implicit_decoder}. With a condition, such as a single image~\cite{xu2019disn,DBLP:conf/cvpr/ChibaneAP20,Genova:2019:LST,seqxy2seqzeccv2020paper} or a learnable latent code~\cite{Park_2019_CVPR}, neural networks can be trained as an implicit function to model various shapes. We can also leverage point clouds as conditions~\cite{Williams_2019_CVPR,liu2020meshing,Mi_2020_CVPR,Genova:2019:LST} to learn implicit functions, and then leverage the marching cubes algorithm~\cite{Lorensen87marchingcubes} to reconstruct surfaces~\cite{jia2020learning,ErlerEtAl:Points2Surf:ECCV:2020}. To capture more detailed geometry, implicit functions are defined in local regions which are covered by voxel grids~\cite{jiang2020lig,DBLP:conf/eccv/ChabraLISSLN20,Peng2020ECCV,DBLP:journals/corr/abs-2105-02788,takikawa2021nglod,Liu2021MLS,tang2021sign}, patches~\cite{Tretschk2020PatchNets}, 3D Gaussian functions~\cite{Genova_2020_CVPR}, learnable codes~\cite{DBLP:conf/cvpr/LiWLSH22,Boulch_2022_CVPR}.

\noindent\textbf{Learning from Multi-view Images. } \js{Some recent works learn implicit functions from 2D supervision, such as multiple images. The basic idea is to leverage various differentiable renderers~\cite{sitzmann2019srns,DIST2019SDFRcvpr,Jiang2019SDFDiffDRcvpr,DBLP:journals/cgf/WuS20,Volumetric2019SDFRcvpr,lin2020sdfsrn} to render the learned implicit functions into images, so that we can obtain the error between rendered images and ground truth images. Neural volume rendering was introduced to capture the geometry and color simultaneously~\cite{mildenhall2020nerf,yariv2020multiview,yariv2021volume,geoneusfu,neuslingjie,Yu2022MonoSDF,yiqunhfSDF,huang2023neusurf}. By sampling rays emitted from pixels into the field, UNISURF \cite{Oechsle2021ICCV} and NeuS \cite{neuslingjie} employ a modified rendering procedure to transform occupancy and signed distance fields, along with radiance, into pixel colors. Subsequent methods enhance the accuracy of implicit functions by leveraging multi-view consistency \cite{Darmon_2022_CVPR, geoneusfu} and incorporating additional priors such as depth \cite{Yu2022MonoSDF} and normals \cite{wang2022neuris, zhu2021adafit}.}

\noindent\textbf{Learning from 3D Point Clouds. }Some methods were proposed to learn implicit functions from point clouds without 3D ground truth. These methods leverage additional constraints~\cite{DBLP:conf/icml/GroppYHAL20,Atzmon_2020_CVPR,zhou2023uni3d,zhao2020signagnostic,atzmon2020sald,wen20223d,Zhou2023VP2P,DBLP:journals/corr/abs-2106-10811,zhou20223d,yifan2020isopoints,ben2021digs,zhou2022-3DOAE}, gradients~\cite{Zhizhong2021icml,chibane2020neural}, differentiable poisson solver~\cite{Peng2021SAP} or specially designed priors~\cite{DBLP:conf/cvpr/MaLH22,DBLP:conf/cvpr/MaLZH22} to learn signed~\cite{Zhizhong2021icml,DBLP:conf/icml/GroppYHAL20,Atzmon_2020_CVPR,zhao2020signagnostic,atzmon2020sald,chaompi2022,VisCovolume,jin2023multi} or unsigned distance fields~\cite{chibane2020neural,Zhou2022CAP-UDF,udiff,zhou2024cap}. One issue here is that they usually assume the point clouds are clean, which limits their performance in real applications due to the noise. Our method falls into this category, but we can resolve this problem using statistical reasoning via noise to noise mapping.

\noindent\textbf{Deep Learning based Point Cloud Denoising. }PointCleanNet~\cite{DBLP:journals/cgf/RakotosaonaBGMO20} was introduced to remove outliers and reduce noise from point clouds using a data-driven strategy. Graph convolution was also leveraged to reduce the noise based on dynamically constructed neighborhood graphs~\cite{DBLP:conf/eccv/PistilliFVM20}. Without supervision, TotalDenoising~\cite{DBLP:conf/iccv/Casajus0R19} inherits the same idea as Noise2Noise~\cite{DBLP:conf/icml/LehtinenMHLKAA18}. It leveraged a spatial prior term that can work for unordered point clouds. More recently, downsample-upsample architecture~\cite{DBLP:conf/mm/LuoH20} and gradient fields~\cite{luo2021score,ShapeGF} were leveraged to reduce noise. We were inspired by the idea of Noise2Noise~\cite{DBLP:conf/icml/LehtinenMHLKAA18}, our contribution lies in our finding that we can still leverage statistical reasoning among multiple noisy point clouds with specially designed losses \js{even when there is} no spatial correspondence among points on different observations like the one among pixels, which is totally different from TotalDenoising~\cite{DBLP:conf/iccv/Casajus0R19}.

\section{Method}
\label{sec:method}

\noindent\textbf{Overview. } Given $N$ corrupted observations $S=\{\bm{S}_i|i\in[1,N],N\ge 1\}$ of an uncorrupted 3D shape or scene $\bm{Z}$, we aim to learn SDFs $f$ of $\bm{Z}$ from $S$ without ground truth signed distances, point normals, or clean point clouds. Here, $\bm{S}_i$ is a noisy point cloud. \js{The signed distance function $f$ predicts a signed distance $d$ for an arbitrary query location $\bm{q}\in \mathbb{R}^{1\times 3}$ around $\bm{Z}$, such that $d=f(\bm{q})$. We train a neural network with parameters $\bm{\theta}$ to learn $f$, which we denote as $f_{\bm{\theta}}$. After training, we can leverage the learned $f_{\bm{\theta}}$ for surface reconstruction, point cloud denoising, and point cloud upsampling.}

\begin{figure}[tb]
  \centering
   \includegraphics[width=\linewidth]{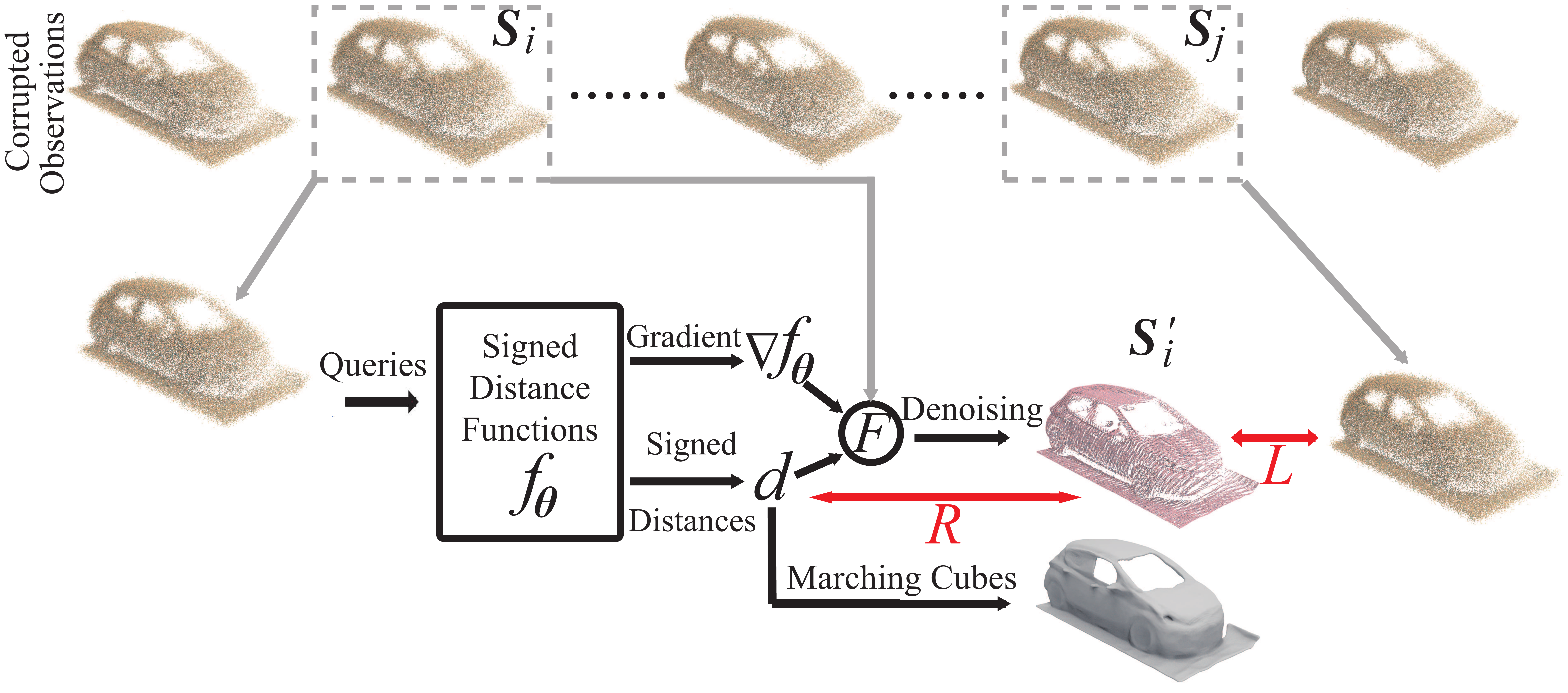}
\caption{\label{fig:OverviewGeneral}\js{Given corrupted observations captured by a Lidar system per second, we learn a SDF without supervision or normals.}}
\end{figure}

Our key idea of statistical reasoning is demonstrated in Fig.~\ref{fig:OverviewGeneral}. Using a noisy point cloud $\bm{S}_i$ as input, our network aims to learn SDFs $f_{\bm{\theta}}$ via learning a noise to noise mapping from $\bm{S}_i$ to another noisy point cloud $\bm{S}_j$, where $\bm{S}_j$ is also randomly selected from the corrupted observation set $S$ and $j\in[1,N]$. Our loss not only minimizes the distance between the denoised point cloud $\bm{S}_i'$ and $\bm{S}_j$ using a metric $L$ but also constrains the learned SDFs $f_{\bm{\theta}}$ to be correct using a geometric consistency regularization $R$. A denoising function $F$ conducts point cloud denoising using signed distances $d$ and gradients $\nabla f_{\bm{\theta}}$ from $f_{\bm{\theta}}$.

\noindent\textbf{Reducing Noise. }A common strategy for estimating the uncorrupted data from its noise corrupted observations is to find a target that has the smallest average deviation from measurements according to some loss function $L$. The data could be a scalar, a 2D image or a 3D point cloud etc.. Here, to reduce noise on point clouds, we aim to find the uncorrupted point cloud $\bm{S}'$ from its corrupted observations $\bm{S}\in S$ below,
\begin{equation}
\label{eq:1}
\begin{aligned}
\argmin_{\bm{S}'} \mathbb{E}_{\bm{S}} \{L(\bm{S}',\bm{S})\}.
\end{aligned}
\end{equation}
As a conclusion of Noise2Noise~\cite{DBLP:conf/icml/LehtinenMHLKAA18} for 2D image denoising, we can learn a denoising function $F$ by pushing a denoised image $F(\bm{x})$ to be similar to as many corrupted observations $\bm{y}$ as possible, where both $\bm{x}$ and $\bm{y}$ are corrupted observations. This is an appealing conclusion since we do not need the expensive pairs of the corrupted inputs and clean targets to learn the denoising function $F$.

We want to leverage this conclusion to learn to reduce noise without requiring clean point clouds. So we transform Eq.~(\ref{eq:1}) into an equation with a denoising function $F$,
\begin{equation}
\label{eq:2}
\begin{aligned}
\argmin_{F} \sum_{\bm{S}_i\in S}\sum_{\bm{S}_j\in S} L(F(\bm{S}_i),\bm{S}_j).
\end{aligned}
\end{equation}
One issue we are facing is that the conclusion of Noise2Noise may not work for 3D point clouds, due to the irregular and unordered characteristics of point clouds. For 2D images, multiple corrupted observations have the pixel correspondence. This results in an assumption that all noisy observations at the same pixel location are random realizations of a distribution around a clean pixel value. However, this assumption is invalid for point clouds. This is also the reason why TotalDenoising~\cite{DBLP:conf/iccv/Casajus0R19} does not think Eq.~(\ref{eq:1}) can work for point cloud denoising, since the noise in 3D point clouds is total. Differently, our finding is in opposite direction. We think we can still leverage Eq.~(\ref{eq:1}) to reduce noise in 3D point clouds, and the key is how to define the distance metric $L$, which is regarded as one of our contributions.

\begin{figure}[b]
  \centering
   \includegraphics[width=0.85\linewidth]{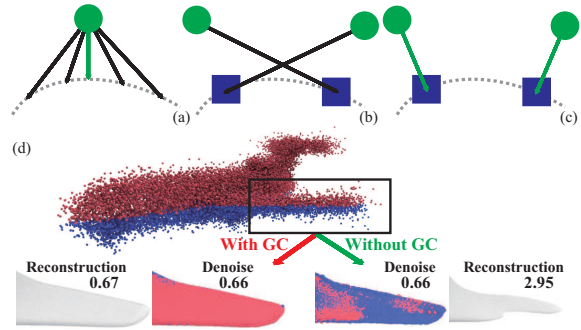}
  
\caption{\label{fig:Second}(a) Multiple paths (arrows) to pull a noise (green point) onto surface (dashed curve) but only one is the shortest (green arrows). (b) The incorrect paths (black arrows) to pull noises onto surface. (c) The expected paths (green arrows) to pull noises to points (blue square) on surface. (d) The effect of Geometric Consistency (GC).}

\end{figure}

Another issue that we are facing is how we can learn SDFs $f_{\bm{\theta}}$ via point cloud denoising in Eq.~(\ref{eq:2}). Our solution is to leverage $f_{\bm{\theta}}$ to define the denoising function $F$. This enables to conduct the learning of SDFs and point cloud denoising at the same time. Next, we will elaborate on our solutions to the aforementioned two issues.

\noindent\textbf{Denoising Function $F$. }The denoising function $F$ aims to produce a denoised point cloud $\bm{S}'$ from a noisy point cloud $\bm{S}$, so $\bm{S}'=F(\bm{S})$.

To learn SDFs $f_{\bm{\theta}}$ of $\bm{S}$, we want the denoising procedure can also perceive the signed distance fields around $\bm{S}$. The essence of denoising is to move points floating off the surface of an object onto the surface. As shown in Fig.~\ref{fig:Second} (a), there are many potential paths to achieve this, but only one path is the shortest to the surface. If we leverage this shortest path to denoise point cloud $\bm{S}$, we could involve the SDFs $f_{\bm{\theta}}$ to define the denoising function $F$, since $f_{\bm{\theta}}$ can determine the shortest path.

\js{Here, inspired by the idea of NeuralPull~\cite{Zhizhong2021icml}, we also leverage the signed distance $d=f_{\bm{\theta}}(\bm{q})$ and the gradient $\nabla f_{\bm{\theta}}(\bm{q})$ to pull an arbitrary point $\bm{q}$ on the noisy point cloud $\bm{S}$ onto the surface. So we define the denoising function $F$ below,}
\begin{equation}
\label{eq:3}
\begin{aligned}
F(\bm{q},f_{\bm{\theta}})=\bm{q}-d\times\nabla f_{\bm{\theta}}(\bm{q})/||\nabla f_{\bm{\theta}}(\bm{q})||_2.
\end{aligned}
\end{equation}
With Eq.~(\ref{eq:3}), we can pull all points on the noisy point cloud $\bm{S}$ onto the surface, which results in a point cloud $\bm{S}'=F(\bm{S},f_{\bm{\theta}})$. But one issue remaining is how to constrain $\bm{S}'$ to converge to the uncorrupted surface.

\noindent\textbf{Distance Metric $L$. }We investigate the distance metric $L$ so that we can constrain $\bm{S}'$ to reveal the uncorrupted surface by a statistical reasoning among the corrupted observations $S=\{\bm{S}_i\}$ using Eq.~(\ref{eq:2}). Our investigation conclusion is summarized in the following Theorem.

\noindent\textbf{Theorem 1. }\textit{Assume there was a clean point cloud $\bm{G}$ which is corrupted into observations $S=\{\bm{S}_i\}$ by sampling a noise around each point of $\bm{G}$. If we leverage EMD as the distance metric $L$ defined in Eq.~(\ref{eq:4-1}), and learn a point cloud $\bm{G}'$ by minimizing the EMD between $\bm{G}'$ and each observation in $S$, i.e., $\min_{\bm{G}'}\sum_{\bm{S}_i\in S}L(\bm{G}',\bm{S}_i)$, then $\bm{G}'$ converges to the clean point cloud $\bm{G}$, i.e., $L(\bm{G},\bm{G}')=0$.}
\begin{equation}
\label{eq:4-1}
\begin{aligned}
L(\bm{G},\bm{G}')=\min_{\phi:\bm{G}\to\bm{G}'}\sum_{\bm{g}\in\bm{G}}||\bm{g}-\phi(\bm{g})\|_2.
\end{aligned}
\end{equation}

\begin{figure}[tb]
  \centering
   \includegraphics[width=\linewidth]{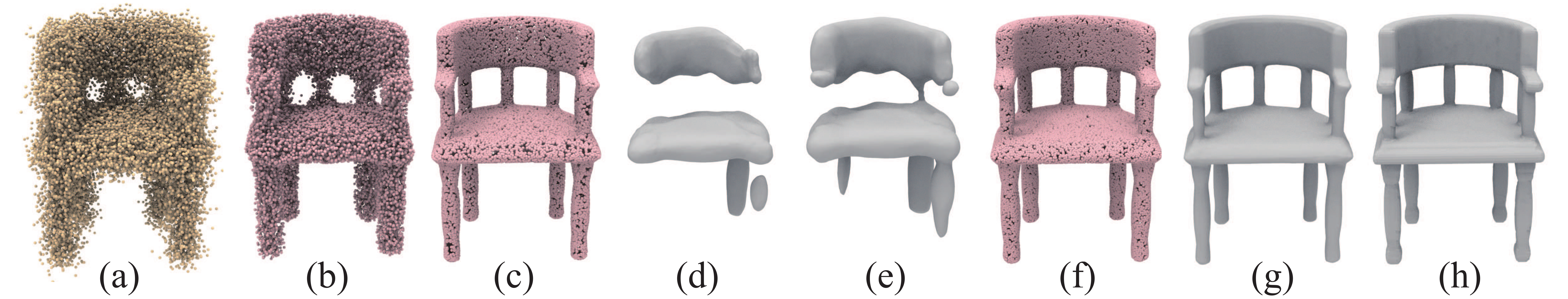}
  
\caption{\label{fig:CDEMD}The comparison with CD and EMD as the distance metric $L$ from in (b) to (e). The effect of geometric regularization in (f) and (g). (a) is noisy point cloud, (h) is the ground truth.}
\end{figure}

We prove Theorem 1 in Sec.A of the supplementary materials. We believe the one-to-one correspondence $\phi$ found in the calculation of EMD in Eq.~(\ref{eq:4-1}) plays a big role in the statistical reasoning for denoising. This is very similar to the pixel correspondence among noisy images in Noise2Noise although point clouds are irregular, unordered and have no spatial correspondence among points on different observations. We highlight this by comparing the point cloud $\bm{G}'$ optimized with EMD and Chamfer Distance (CD) as $L$ based on the same observation set $S$ in Fig.~\ref{fig:CDEMD}. Given noisy point clouds $\bm{S}_i$ like in Fig.~\ref{fig:CDEMD} (a), Fig.~\ref{fig:CDEMD} (b) demonstrates that the point cloud $\bm{G}'$ optimized with CD is still noisy, while the one optimized with EMD in Fig.~\ref{fig:CDEMD} (c) is very clean.

According to this theorem, we can learn the denoising function $F$ using Eq.~(\ref{eq:2}). $F$ produces the denoised point cloud $\bm{S}_i'=F(\bm{S}_i,f_{\bm{\theta}})$ using EMD as the distance metric $L$. This also leads to one term in our loss function below,
\begin{equation}
\label{eq:5}
\begin{aligned}
\min_{\theta} \sum_{\bm{S}_i\in S}\sum_{\bm{S}_j\in S} L_{\mathrm{EMD}}(F(\bm{S}_i,f_{\bm{\theta}}),\bm{S}_j).
\end{aligned}
\end{equation}

\noindent\textbf{Geometric Consistency. }Although the term in Eq.~(\ref{eq:5}) can work for point cloud denoising well, as shown in Fig.~\ref{fig:CDEMD} (c), we found that the SDFs $f_{\theta}$ may not describe a correct signed distance field. With $f_{\theta}$ either learned with CD or EMD, the surfaces reconstructed using marching cubes algorithms~\cite{Lorensen87marchingcubes} in Fig.~\ref{fig:CDEMD} (d) and (e) are poor. This is because Eq.~(\ref{eq:5}) only constrains that points on the noisy point cloud should arrive onto the surface but there are no constraints on the paths to be the shortest. This is caused by the unawareness of the true surface which however is required as the ground truth by NeuralPull~\cite{Zhizhong2021icml}. The issue is further demonstrated in Fig.~\ref{fig:Second}, one situation that may happen is shown in Fig.~\ref{fig:Second} (b). With the wrong signed distances $f_{\bm{\theta}}$ and gradient $\nabla f_{\bm{\theta}}$, noises can also get pulled onto the surface, which results in a denoised point cloud with zero EMD distance to the clean point clouds. This is much different from the correct signed distance field that we expected in Fig.~\ref{fig:Second} (c).

To resolve this issue, we introduce a geometric consistency to constrain $f_{\theta}$ to be correct. Our insight here is that, for an arbitrary query $\bm{q}$ around a noisy point cloud $\bm{S}_i$, the shortest distance between $\bm{q}$ and the surface can be either predicted by the SDFs $f_{\theta}$ or calculated based on the denoised point cloud $\bm{S}_i'=F(\bm{S}_i,f_{\theta})$, both of which should be consistent to each other. \js{Therefore, the absolute value $|f_{\theta}(\bm{q})|$ of the signed distance predicted at $\bm{q}$ should equal to the minimum distance between $\bm{q}$ and the denoised point cloud $\bm{S}_i'=F(\bm{S}_i,f_{\theta})$.} Since the point density of $\bm{S}_i'$ may slightly affect the consistency, we leverage an inequality to describe the geometric consistency,
\begin{equation}
\label{eq:6}
\begin{aligned}
|f_{\theta}(\bm{q})| \le \min_{\bm{q}'\in F(\bm{S}_i,f_{\theta})} ||\bm{q}-\bm{q}'||_2.
\end{aligned}
\end{equation}
The geometric consistency is further illustrated in Fig.~\ref{fig:Second} (d). Noisy points above/below the wing can be correctly pulled onto the upper/lower surface without crossing the wing using the geometric consistency. It achieves the same denoising performance, and leads to a much more accurate SDF for surface reconstruction than the one without the geometric consistency.

\noindent\textbf{Loss Function. }With the geometric consistency, we can penalize the incorrect signed distance field shown in Fig.~\ref{fig:Second} (b) while encouraging the correct one in Fig.~\ref{fig:Second} (c). So, we leverage the geometric consistency as a regularization term $R$, which leads to our objective function below by combining Eq.~(\ref{eq:5}) and Eq.~(\ref{eq:6}),
\begin{equation}
\label{eq:7}
\begin{aligned}
\min_{\theta} \sum_{\bm{S}_i\in S}(\sum_{\bm{S}_j\in S} L(F(\bm{S}_i,f_{\theta}),\bm{S}_j)+\frac{\lambda }{|\bm{S}_i|}\sum_{\bm{q}\in\bm{S}_i}R(E),
\end{aligned}
\end{equation}
\noindent where $|\bm{S}_i|$ is the number of $\bm{q}$ on $\bm{S}_i$, $E$ is the difference defined as $(|f_{\theta}(\bm{q})|-\min_{\bm{q}'\in F(\bm{S}_i,f_{\theta})} ||\bm{q}-\bm{q}'||_2)$, $\lambda$ is a balance weight, and $R(E)=max(0,E)$. The effect of the geometric consistency is demonstrated in Fig.~\ref{fig:CDEMD} (f) and (g). The denoised point cloud in Fig.~\ref{fig:CDEMD} (f) shows points that are more uniformly distributed, compared with the one obtained without the geometric consistency in Fig.~\ref{fig:CDEMD} (c). More importantly, we can learn correct SDFs $f_{\theta}$ to reconstruct plausible surface in Fig.~\ref{fig:CDEMD} (g), compared to the one obtained without the geometric consistency in Fig.~\ref{fig:CDEMD} (e) and the ground truth in Fig.~\ref{fig:CDEMD} (h).

\noindent\textbf{More Details. }We sample more queries around the input noisy point cloud $\bm{S}_i$ using the method introduced in NeuralPull~\cite{Zhizhong2021icml}. We randomly sample a batch of $B$ queries as input, and also randomly sample the same number of points from another noisy point cloud $\bm{S}_j$ as target. Using batches enables us to process large scale point clouds,  makes it possible to leverage noisy point clouds with different point numbers even we use EMD as the distance metric $L$, and more importantly, does not affect the performance. 

\begin{figure*}[htb]
  \centering
   \includegraphics[width=\linewidth]{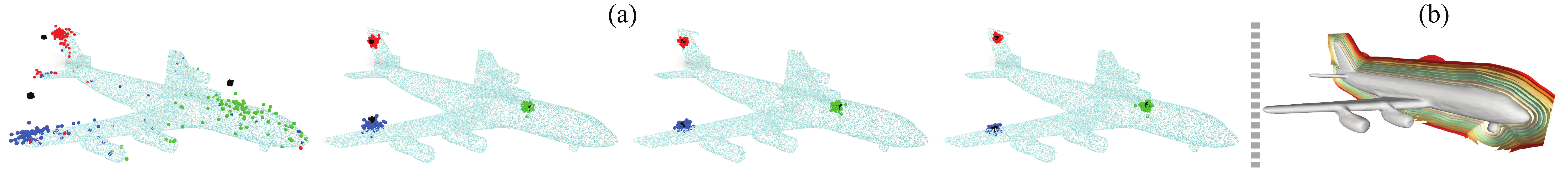}
  
\caption{\label{fig:NoiseOpt}(a) Visualization of optimization in $4$ epochs via noise to noise mapping. $3$ queries (black cubes) sampled from one noisy point cloud get pulled onto the surface. For each query, we minimize its distance to all targets (in the same color) matched from another noisy point cloud by the mapping $\phi$ in metric $L$. More details can be found in our video at the project page. (b) Surface reconstruction and multiple level-sets.}
\end{figure*}

We visualize the optimization process in $4$ epochs in Fig.~\ref{fig:NoiseOpt} (a). We show how the $3$ queries (black cubes) get pulled progressively onto the surface (Cyan). For each query, we also show its corresponding target in each one of $100$ batches in the same color (red, green, blue), and each target is established by the mapping $\phi$ in the metric $L$. The essence of \js{statistical} reasoning in each epoch is that each query will be pulled to the average point of all targets from all batches since the distance between the query and each target should be minimized. Although the targets are found all over the shape in the first epoch, the targets surround the query more tightly as the query gets pulled to the surface in the following epochs. This makes queries get pulled onto the surface which results in an accurate SDF visualized in the surface reconstruction and level-sets in Fig.~\ref{fig:NoiseOpt} (b).

\noindent\textbf{One Noisy Point Cloud. }Although we prove Theorem $1$ based on multiple noisy point clouds ($N>1$), we surprisingly found that our method can also work well when only one noisy point cloud ($N=1$) is available. Specifically, we regard the queries sampled around the noisy point cloud $\bm{S}_i$ as input and regard $\bm{S}_i$ as target. We believe the reason why $N=1$ works is that the knowledge learned via statistical reasoning in the batch based training can be well generalized to various regions. We will report our results learned from multiple or one noisy point clouds in experiments.

\begin{figure}[tb]
 
  \centering
   \includegraphics[width=\linewidth]{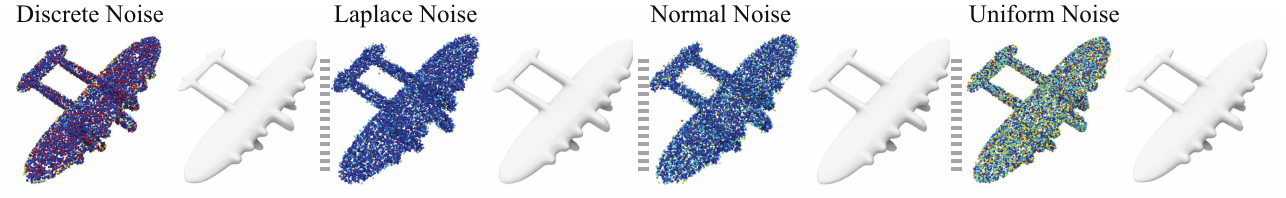}
  
\caption{\label{fig:assumption}Reconstruction with different kinds of noises.}

\end{figure}

\noindent\textbf{Noise Types. }We work well with different types of noises in Fig.~\ref{fig:assumption}. We use zero-mean noises in our proof of Theorem 1, but we find we work well with unknown noises in real scans in experiments. In evaluations, we also use the same type of noises in benchmarks for fair comparisons.

\section{\md{Fast Learning of Implicit Representations}}
\label{sec:ngp}
\noindent\textbf{Review Instant-NGP.}
The deep coordinate-based MLPs are proven to be slow in learning implicit functions (e.g. NeRF and SDF). To overcome this issue, Instant-NGP \cite{mueller2022instant} leverages a multi-resolution hash encoding to learn radiance fields. Specifically, Instant-NGP divides the bounded space of a 3D shape into multiple voxel grids at different resolutions. The voxel grids at each resolution are then mapped to a hash table with learnable feature vectors. 

\begin{figure}[tb]
  \centering
   \includegraphics[width=\linewidth]{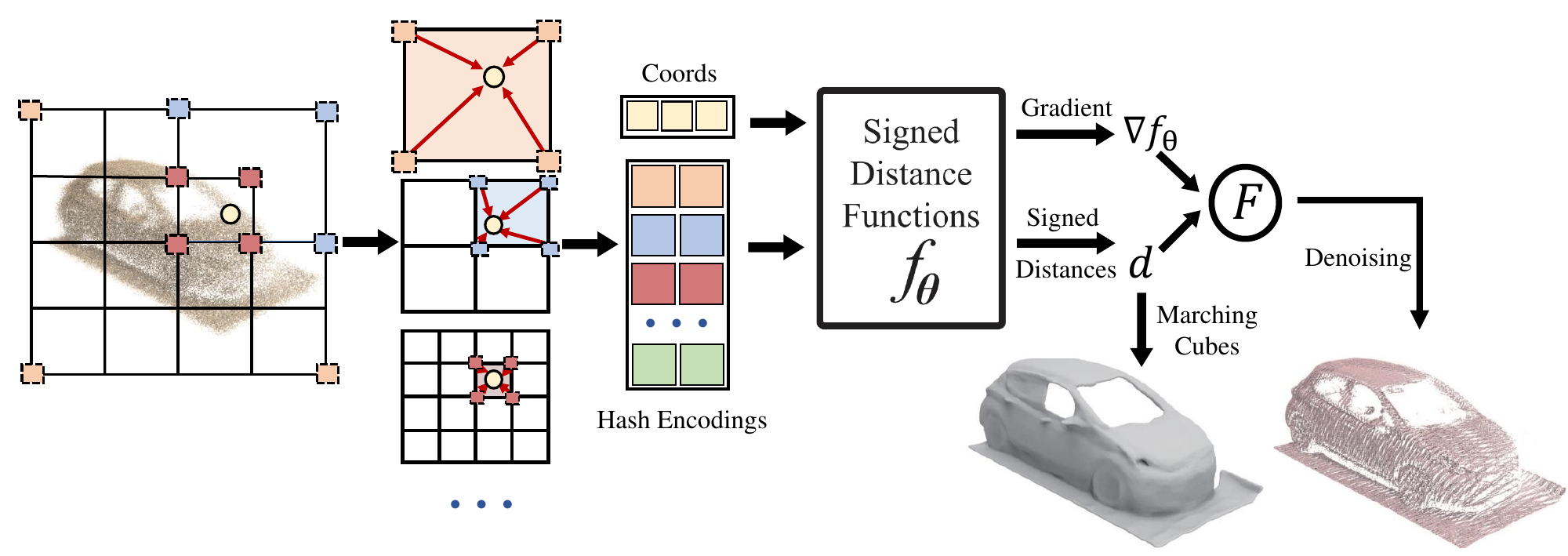}
  
\caption{\label{fig:ngp} \js{Overview of our fast learning framework. We leverage the multi-resolution hash encoding as the efficient representation for fast learning SDFs from noisy point clouds.}}
\end{figure}

\begin{figure*}[htb]
  \centering

   \includegraphics[width=\textwidth]{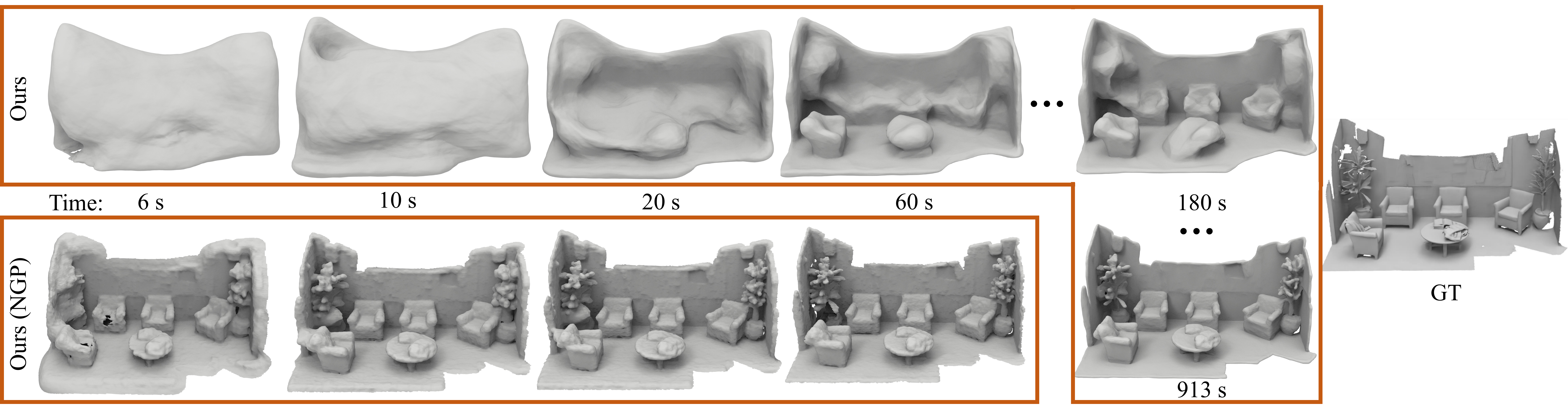}

\caption{\label{fig:ngp_results} Comparisons on the reconstructions at different optimization time. The timing results do not include the time for Marching Cubes, and are reported under one single RTX3090 GPU.}
\end{figure*}

\noindent\textbf{Fast Learning of SDFs from Noisy Point Clouds.}
We propose a fast learning schema to learn signed distance fields from noisy point clouds by leveraging the multi-resolution hash encoding \cite{mueller2022instant} as an efficient representation. The overview of our approach is shown in Fig. \ref{fig:ngp}. We first initialize a hash encoding with $L$ levels of resolution, where each level $l = 1,...,L$ has a hash table $T_l$. Given a point $ \bm{q} \in \mathcal{R}^{1\times 3}$ from one noisy point cloud $\bm{S}$, we obtain its feature $h_l(\bm{q})$ at level $l$ using trilinear interpolating on the hash encodings of the eight surrounding corner grids $g_i$ queried from the hash table $T_l$. The hash encoded feature $\{h_l(\bm{q})\}_{l=1}^L$ of $\bm{q}$ at all $L$ levels, as well as the coordinate of $\bm{q}$ are then concatenated as the input to the MLP $f_{\bm{\theta}}$ to decode the signed distance value $d$. The similar denoising function as Eq. (\ref{eq:3}) is than leveraged for moving points floating off the surface of 3D shape onto the surface as:

\begin{equation}
\label{eq:move2}
\begin{aligned}
F_{\mathrm{Fast}}(\bm{q}&,\{h_l(n)\}_{l=1}^L,f_{\bm{\theta}})=\bm{q}-\\
&d\times\nabla f_{\bm{\theta}}(\bm{q},\{h_l(n)\}_{l=1}^L)/||\nabla f_{\bm{\theta}}(\bm{q},\{h_l(n)\}_{l=1}^L)||_2.
\end{aligned}
\end{equation}
The condition here is the hash encoded feature $\{h_l(\bm{q})\}_{l=1}^L$ of $\bm{q}$ instead of the feature of entire shape $\bm{Z}$.
We then optimize both the feature vectors in the multi-resolution hash encodings and the parameter $\bm{\theta}$ of MLP decoder $f_{\bm{\theta}}$ by minimizing the designed objective function in Eq. (\ref{eq:7}) which combines the EMD loss in Eq. (\ref{eq:5}) and the geometric consistency loss in Eq. (\ref{eq:6}). We provide the visualization of reconstructions at different optimization time with or without our hash encodings in Fig. \ref{fig:ngp_results}. The results show that our fast learning schema reduces the training time from 15 minutes to one minute with comparable surface qualities, and is able to produce a visual appearing reconstruction within 6 to 10 seconds. All timing results are reported with one single RTX3090 GPU.

\noindent\textbf{More Analyses and Details.} Different from the original version of our method \cite{BaoruiNoise2NoiseMapping} which learns deep coordinate-based MLPs for geometry representation, the learning of multi-resolution hash encodings is more difficult to stabilize and requires some other constraints besides the EMD loss in Eq. (\ref{eq:5}). 
The main problem lies in the instability of the Earth Moving Distance (EMD) loss during optimizations in regions far away from the surface, as establishing a one-to-one correspondence between queries far away from the surface and the point cloud $S$ is challenging. This was not a significant issue in the previous MLP-based version \cite{BaoruiNoise2NoiseMapping}, which only focused on the surface's nearby space. However, optimizing the far-space regions is crucial for multi-resolution hash encodings, as it ensures that the hash grids in all spatial spaces are well-optimized, leading to refined representations. To solve this issue, we sample queries $\{Q = {q_i}, i \in [1, N_{q}]\}$ at all spacial space and leverage Neural-Pull \cite{ma2020neuralpull} loss to regularize the gradients and SDFs (especially in areas that is far away from the surface) by pulling each query ${q}_i$ to its nearest point ${\bm{n'}}_i$ in $S$, formulated as:

\begin{equation}
    \label{eq:pullloss}
    L_{\mathrm{pull}} = \frac{1}{N_{q}} \sum_{i = 1}^{N_{q}} {||q_i-{n'}_i||_2^2}.
\end{equation}

The eikonal term \cite{gropp2020implicit} is also used to regularize the gradients of SDFs, formulated as:

\begin{equation}
    \label{eq:eikonal}
    L_{\mathrm{reg}}(Q, \bm{c}) = \frac{1}{N_{q}} \sum_{i = 1}^{N_{q}}{(|\nabla f_{\bm{\theta}}({q}_i, c)|-1)^2}
\end{equation}

The finally loss function is formulated as:

\begin{equation}
    L_{\mathrm{Fast}} = L_\mathrm{EMD} + \lambda_1 L_\mathrm{pull} + \lambda_2 L_\mathrm{reg},
\end{equation}
where $\lambda_2$ is set to 0.001. $\lambda_1$ is set to decrease from 1 to 0 during the first 1000 iterations and is omitted for the rest iterations, where we train our fast learning framework for 10000 iterations in total. We only leverage $L_\mathrm{pull}$ to stable the optimizations in the areas far away from the surface and omit it after a fine initialized fields is achieved, since it will lead to large errors to pull the queries to the noisy inputs in the areas near to the true surface.

\begin{figure*}[tb]
  \centering

   \includegraphics[width=\linewidth]{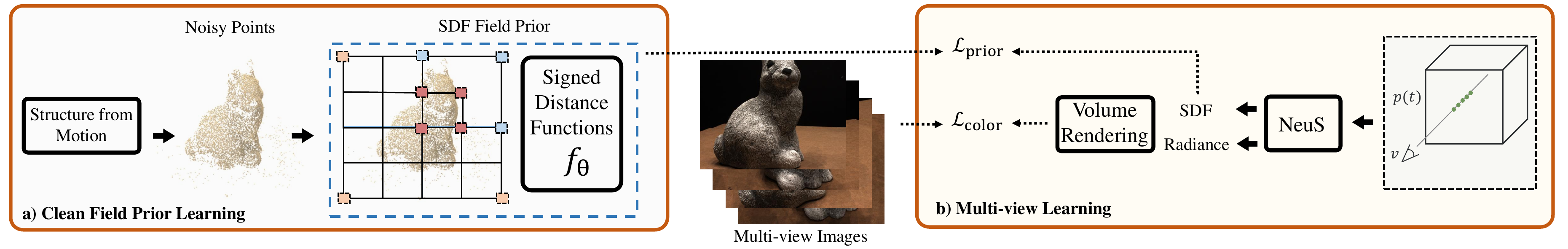}

\caption{\label{fig:prior} Overview of our multi-view reconstruction framework. (a) We extract a noisy and sparse point cloud from multi-view images with SfM and learn a clean SDF field prior from it with our fast learning denoising approach. (b) We guide the multi-view learning (NeuS) with the learned clean field prior by introducing $\mathcal{L}_{\rm{prior}}$.}
\end{figure*}

\section{\md{Learning SDF Priors for Multi-View Reconstruction}}
\noindent\textbf{Review NeuS.} 
We extend our noise to noise mapping to learn SDF priors for multi-view reconstruction. We use NeuS as our baseline. NeuS \cite{neuslingjie} is a recent work which focuses on reconstructing surfaces from a set of calibrated images $\{\mathcal{I}_k\}$ of a 3D object. To achieve this, NeuS implicitly learns a signed distance function $f_{mv}(x):\mathbb{R}^3 \rightarrow \mathbb{R}$ and a radiance function $c(x,v):\mathbb{R}^3 \times \mathbb{S}^2$, where $x \in \mathbb{R}^3$ is a 3D location and $v \in \mathbb{S}^2$ is a view direction. The surfaces of the 3D object are implicitly represented as the zero-level set of $f_{mv}$ and the appearances are encoded in $c$. To optimize the two functions, NeuS leverages volume rendering to render novel images from $f_{mv}$ and $c$ and minimize the differences between the rendered images and input images. Specifically, given a pixel of an image, we sample $h$ queries along the ray emitted from this pixel as $\{p(t)=o+tv|t=0,1...h\}$, where $o$ is the camera center and $v$ is the view direction. The color of this pixel is then obtained by accumulating the SDF-based densities and colors at the sampled queries.  

\begin{table*}[ht]
\setlength{\tabcolsep}{3mm}
\centering
\resizebox{0.9\linewidth}{!}{
    \begin{tabular}{c|c|cccccc|cccccc}  %
     \toprule
        \multicolumn{2}{c|}{Point Number}&\multicolumn{6}{c|}{10K(Sparse)}&\multicolumn{6}{c}{50K(Dense)}  \\
        \hline
        \multicolumn{2}{c|}{Noise}&\multicolumn{2}{c}{1\%}&\multicolumn{2}{c}{2\%}&\multicolumn{2}{c|}{3\%}&\multicolumn{2}{c}{1\%}&\multicolumn{2}{c}{2\%}&\multicolumn{2}{c}{3\%}  \\
        &Model&CD&P2M&CD&P2M&CD&P2M&CD&P2M&CD&P2M&CD&P2M\\
    \hline
        \multirow{9}{*}{\rotatebox{90}{PU}}&\multicolumn{1}{l|}{Bilateral}&3.646&1.342&5.007&2.018&6.998&3.557&0.877&0.234&2.376&1.389&6.304&4.730 \\
        &\multicolumn{1}{l|}{Jet}&2.712&0.613&4.155&1.347&6.262&2.921&0.851&0.207&2.432&1.403&5.788&4.267\\
        &\multicolumn{1}{l|}{MRPCA}&2.972&0.922&3.728&1.117&5.009&1.963&0.669&\textbf{0.099}&2.008&1.003&5.775&4.081\\
        &\multicolumn{1}{l|}{GLR}&2.959&1.052&3.773&1.306&4.909&2.114&0.696&0.161&1.587&0.830&3.839&2.707\\
        \cline{2-14}
         &\multicolumn{1}{l|}{PCNet}&3.515&1.148&7.469&3.965&13.067&8.737&1.049&0.346&1.447&0.608&2.289&1.285\\
         &\multicolumn{1}{l|}{GPDNet}&3.780&1.337&8.007&4.426&13.482&9.114&1.913&1.037&5.021&3.736&9.705&7.998\\
         &\multicolumn{1}{l|}{DMR}&4.482&1.722&4.982&2.115&5.892&2.846&1.162&0.469&1.566&0.800&2.632&1.528\\
         &\multicolumn{1}{l|}{SBP}&2.521&0.463&3.686&1.074&4.708&1.942&0.716&0.150&1.288&0.566&1.928&1.041\\
         \cline{2-14}
         &\multicolumn{1}{l|}{TTD-Un}&3.390&0.826&7.251&3.485&13.385&8.740&1.024&0.314&2.722&1.567&7.474&5.729\\
          &\multicolumn{1}{l|}{SBP-Un}&3.107&0.888&4.675&1.829&7.225&3.726&0.918&0.265&2.439&1.411&5.303&3.841\\
         \cline{2-14}
         &\multicolumn{1}{l|}{\textbf{Ours}}&\textbf{1.060}&\textbf{0.241}&\textbf{2.925}&\textbf{1.010}&\textbf{4.221}&\textbf{1.847}
         &\textbf{0.377}&0.155&\textbf{1.029}&\textbf{0.484}&\textbf{1.654}&\textbf{0.972}\\
    \toprule
    \toprule
    \multirow{9}{*}{\rotatebox{90}{PC}}&\multicolumn{1}{l|}{Bilaterall}&4.320&1.351&6.171&1.646&8.295&2.392&1.172&0.198&2.478&0.634&6.077&2.189 \\
        &\multicolumn{1}{l|}{Jet}&3.032&0.830&5.298&1.372&7.650&2.227&1.091&0.180&2.582&0.700&5.787&2.144\\
        &\multicolumn{1}{l|}{MRPCA}&3.323&0.931&4.874&1.178&6.502&1.676&0.966&0.140&2.153&0.478&5.570&1.976\\
        &\multicolumn{1}{l|}{GLR}&3.399&0.956&5.274&1.146&7.249&1.674&0.964&0.134&2.015&0.417&4.488&1.306\\
        \cline{2-14}
         &\multicolumn{1}{l|}{PCNet}&3.849&1.221&8.752&3.043&14.525&5.873&1.293&0.289&1.913&0.505&3.249&1.076\\
         &\multicolumn{1}{l|}{GPDNet}&5.470&1.973&10.006&3.650&15.521&6.353&5.310&1.716&7.709&2.859&11.941&5.130\\
         &\multicolumn{1}{l|}{DMR}&6.602&2.152&7.145&2.237&8.087&2.487&1.566&0.350&2.009&0.485&2.993&0.859\\
         &\multicolumn{1}{l|}{SBP}&3.369&0.830&5.132&1.195&6.776&1.941&1.066&0.177&1.659&0.354&2.494&\textbf{0.657}\\
         \cline{2-14}
         &\multicolumn{1}{l|}{\textbf{Ours}}&\textbf{2.047}&\textbf{0.518}&\textbf{2.056}&\textbf{0.519}&\textbf{5.331}&\textbf{1.935}
         &\textbf{0.426}&\textbf{0.129}&\textbf{1.043}&\textbf{0.316}&\textbf{2.22}&1.096\\
       \toprule
   \end{tabular}%
   }
   \caption{Denoising comparison. L2CD$\times 10^4$ and P2M $\times 10^4$.}
   \label{table:denoising}
\end{table*}

\noindent\textbf{Learn SDF Priors for NeuS.}
The key factor that prevent NeuS \cite{neuslingjie} from producing high-fidelity and artifact-free geometries is the bias in color rendering \cite{yiqunhfSDF, geoneusfu, li2023neuralangelo}. As demonstrated by previous methods \cite{yiqunhfSDF, geoneusfu}, there is a gap between the rendered colors and the real colors of the 3D object surfaces, which leads to geometry inconsistency between the implicit represented surfaces indicated as the zero-level set of field $f_{mv}(*)$ and the true surfaces of the 3D object. To solve this issue, a more precise and adaptive prior is required as the supervision for a better optimization on the signed distance function. 

Specifically, we propose to learn a signed distance field from the point cloud estimated by Structure from Motion (SfM) as a prior for multi-view reconstruction, as  shown in Fig. \ref{fig:prior}(a). The key insight is that the sparse points $P_s$ generated by SfM is an exact geometry information for the 3D object. SfM matches 2D features at the extracted 2D key points to estimate camera poses and compute the 3D point locations through triangulation which guarantees the geometry consistency of $P_s$ in multi-view images. However, the generated $P_s$ often contains noises due to some wrong key point matchings or errors in camera motion estimation. 

We aim to learn a clean and accurate signed distance field from $P_s$ to serve as a prior for multi-view reconstruction, where the noises prevent previous works \cite{ma2020neuralpull, Peng2020ECCV} from predicting an accurate SDF field. To this end, we leverage noise to noise mapping which is robust to noises with different distributions, to learn a SDF prior from the noisy points $P_s$. The learning process keeps the same as Sec. \ref{sec:method} and Sec. \ref{sec:ngp}.

\noindent\textbf{Guide NeuS with SDF Priors.}
With a learned clean field ${f}_{prior}$, we introduce a field constraint to guide the learning of NeuS for improving multi-view reconstruction quality as shown in Fig. \ref{fig:prior}(b). The key factor is to leverage ${f}_{prior}$ as a supervision for the signed distance function $f_{mv}$ to be optimized. Specifically, given a pixel of an image and $h$ queries sampled along the ray emitted from this pixel as $\{p(t)=o+tv|t=0,1...h\}$, we minimize the difference between the predicted signed distance value $f_{mv}(q)$ and the pre-learned signed distance prior ${f}_{prior}(q)$. The designed loss is formulated as:

\begin{equation}
\label{eq:prior}
    L_{\rm{prior}} = \frac{1}{MN} \sum_{i = 1}^{M}{\sum_{j =1}^N}{||f_{mv}(q_{ij})-{f}_{prior}(q_{ij})||_2^2},
\end{equation}
where $M$ is the number of pixels and $N$ is the number of points sampled each pixel. We set the weight of $\mathcal{L}_{\rm{prior}}$ to gradually decrease during the training process for avoiding overfitting the field to the prior. Specifically, we train our framework for 100k iterations and decrease the weight of $\mathcal{L}_{\rm{prior}}$ from 1 to 0 linearly in the first 10k iterations and omit it for the rest iterations.

We further design a novel constraint on the zero level set of $f_{mv}$ with a loss to minimizing the distance errors in the exact surface location. To achieve this, we generate a set of points $\{p^{zls}_i\}_{i=1}^K$ at the zero level set of learned prior ${f}_{prior}$ by moving queries along the gradient directions with a stride of signed distances as Eq. (\ref{eq:move2}). These points $\{p^{zls}_i\}_{i=1}^K$ at the zero level set of the learned prior ${f}_{prior}$ can serve as another prior of the zero level set of $f_{mv}$ with a simple loss to minimize the distance values of $f_{mv}$ at $p^{zls}$, formulated as:
\begin{equation}
\label{eq:zls}
    L_{\rm{zls}} = \frac{1}{K} \sum_{i = 1}^{K}{||f_{mv}(p^{zls}_{k})-0}||_2^2,
\end{equation}

\section{Experiments, Analysis and Applications}

\md{We evaluate our performance in point cloud denoising, upsampling and surface reconstruction from point clouds or multi-view images. We first evaluate our method in point cloud denoising in Sec. \ref{6.1}. Next, we go beyond denoising and evaluate our method in point cloud upsampling in Sec. \ref{6.2}. We then evaluate our method in surface reconstruction from point clouds for shapes in Sec. \ref{6.3} and for scenes in Sec. \ref{6.4}. In Sec. \ref{6.5}, we show the efficiency of our fast learning framework in runtime comparison. We further demonstrate the effectiveness of our method to improve the multi-view reconstruction qualities in Sec. \ref{6.6}. Finally, the ablation studies are shown in Sec. \ref{6.7}.}

\subsection{Point Cloud Denoising}
\label{6.1}

\begin{figure}[tb]
  \centering
   \includegraphics[width=\linewidth]{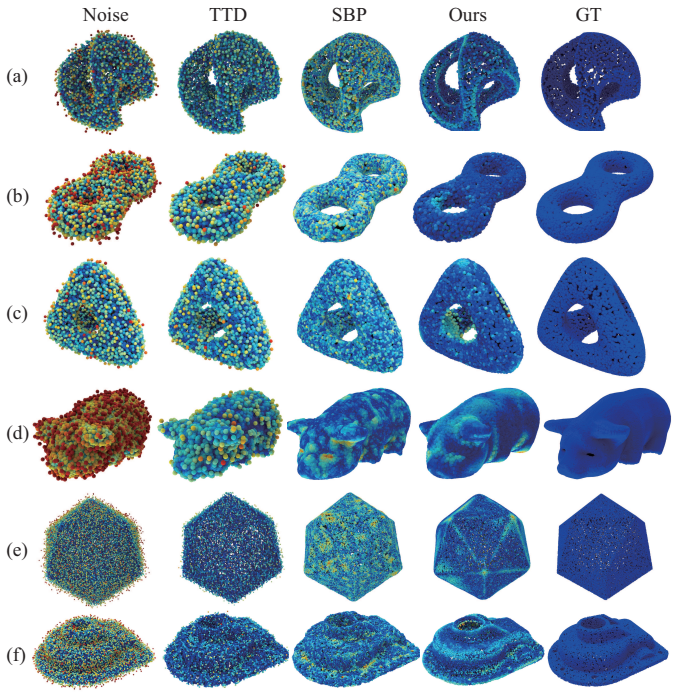}
  
\caption{\label{fig:denoise}Visual comparison in point cloud denoising. Error at each point is shown in color. (a) and (b) $10K$ points with $3\%$ noise. (c) $10K$ points with $2\%$ noise. (d) and (e) $50K$ points with $3\%$ noise. (f) $50K$ points with with $2\%$ noise.}

\end{figure}

\noindent\textbf{Dataset and Metric. }For the fair comparison with the state-of-the-art results, we follow SBP~\cite{luo2021score} to evaluate our method under two benchmarks named as PU and PC that were released by PUNet~\cite{DBLP:conf/cvpr/YuLFCH18} and PointCleanNet~\cite{DBLP:journals/cgf/RakotosaonaBGMO20}. We report our results under 20 shapes in the test set of PU and 10 shapes in the test set of PC. We use Poisson disk to sample $10K$ and $50K$ points from each shape respectively as the ground truth clean point clouds in two different resolutions. The clean point cloud is normalized into the unit sphere. In each resolution, we add Gaussian noise with three standard deviations including $1\%$, $2\%$, $3\%$ to the clean point clouds. We leverage L2 Chamfer Distance (L2CD) and point to mesh distance (P2M) to evaluate the denoising performance.
For each test shape, we generate $N=200$ noisy point clouds to train our method. We sample $B=250$ points in each batch. We report our results and numerical comparison in Tab.~\ref{table:denoising}. The compared methods include Bilateral~\cite{DBLP:journals/tog/FleishmanDC03}, Jet~\cite{DBLP:journals/cagd/CazalsP05}, MRPCA~\cite{DBLP:journals/cgf/MatteiC17}, GLR~\cite{DBLP:journals/tip/ZengCNPY20}, PCNet~\cite{DBLP:journals/cgf/RakotosaonaBGMO20}, GPDNet~\cite{DBLP:conf/eccv/PistilliFVM20}, DMR~\cite{DBLP:conf/mm/LuoH20}, TTD~\cite{DBLP:conf/iccv/Casajus0R19}, and SBP~\cite{luo2021score}. These methods require learned priors and can not directly use multiple observations. The comparison with different conditions indicates that our method significantly outperforms traditional point cloud denoising methods and deep learning based point cloud denoising methods in both supervised and unsupervised (``-Un'') settings. Error map comparison with TTD~\cite{DBLP:conf/iccv/Casajus0R19} and SBP~\cite{luo2021score} in Fig.~\ref{fig:denoise} further demonstrates our state-of-the-art denoising performance.

\subsection{Point Cloud Upsampling}
\label{6.2}
\noindent\textbf{\md{From Denoising to Upsampling. }}
Point cloud upsampling \cite{luo2021score, DBLP:conf/cvpr/YuLFCH18} aims at generating dense and uniformly distributed point sets from a sparse point cloud. We justify that our proposed method is also applicable to point cloud upsampling. Specifically, given a sparse point cloud with $N$ points as input, we perturb it with Gaussian noise independently for $t$ times, resulting in a noisy dense point cloud with $tN$ points, which is then feed to our denoising framework to acquire the final upsampled point cloud.

\noindent\textbf{Dataset and Metric. }
We use the PU dataset mentioned before to evaluate the $f_{\bm{\theta}}$ learned in our denoising experiments in point cloud upsampling. Following SBP~\cite{luo2021score}, we produce an upsampled point cloud with an upsampling rate of 4 from a sparse point cloud by denoising the sparse point cloud with noise. We compare the denoised point cloud and the ground truth, and report L2CD and P2M comparison in Tab.~\ref{table:upsampling}. We compared with PU-Net~\cite{DBLP:conf/cvpr/YuLFCH18} and SBP~\cite{luo2021score}. The comparison demonstrates that our method can perform the statistical reasoning to reveal points on the surface more accurately.
\begin{table}[t]
\centering
\resizebox{1\linewidth}{!}{
    \begin{tabular}{c|ccc|ccc}  %
     \toprule
     \multirow{1}{*}{Points}&\multicolumn{3}{c|}{5K}&\multicolumn{3}{c}{10K} \\
     \hline
     &PU-Net&SBP&\textbf{Ours}&PU-Net&SBP&\textbf{Ours}\\
     \hline
     CD&3.445&1.696&\textbf{0.592}&2.862&1.454&\textbf{0.418}\\
     P2M&1.669&0.295&\textbf{0.156}&1.166&0.181&\textbf{0.155}\\
     \toprule
\end{tabular}%
   }
   \caption{Upsampling comparison. L2CD$\times 10^4$ and P2M $\times 10^4$.}
   \label{table:upsampling}
\end{table}

\subsection{Surface Reconstruction for Shapes}
\label{6.3}

\noindent\textbf{ShapeNet. }We first report our surface reconstruction performance under the test set of 13 classes in ShapeNet~\cite{shapenet2015}. The train and test splits follow COcc~\cite{DBLP:conf/eccv/PengNMP020}. Following IMLS~\cite{Liu2021MLS}, we leverage point clouds with $3000$ points as clean truth, and add Gaussian noise with a standard deviation of 0.005. For each clean point cloud, we generate $N=200$ noisy point clouds with a batch size of $B=3000$. We leverage L1 Chamfer Distance (L1CD), Normal Consistency (NC)~\cite{MeschederNetworks}, and F-score~\cite{Tatarchenko_2019_CVPR} with a threshold of $1\%$ as metrics.

\begin{figure}[tb]
  \centering
   \includegraphics[width=\linewidth]{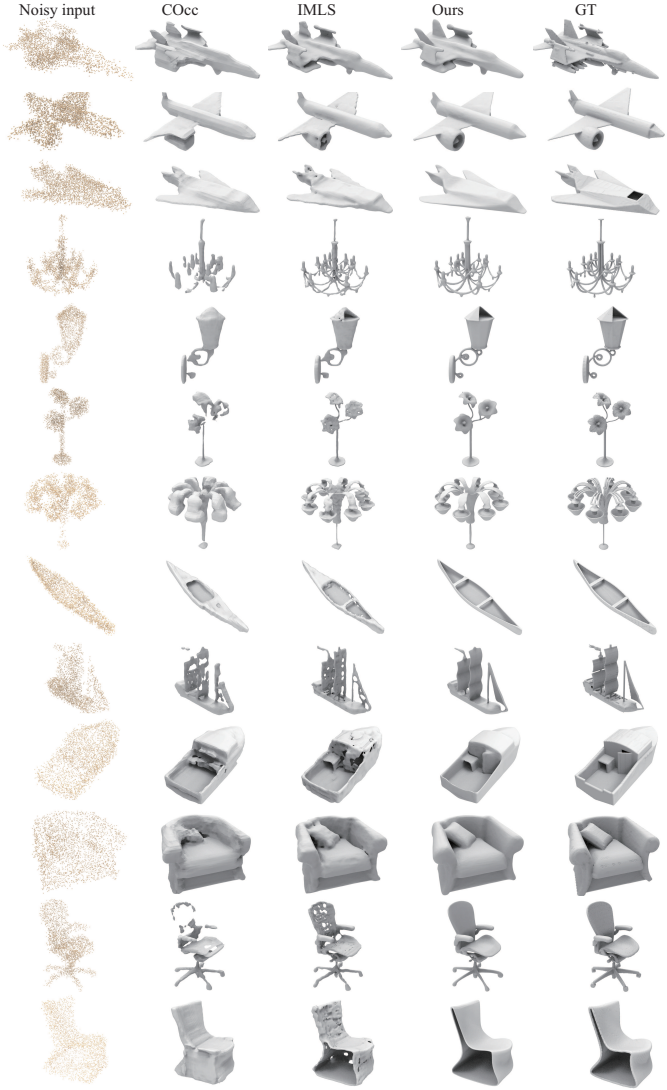}
  
\caption{\label{fig:ShapeNet}Comparison in surface reconstruction under ShapeNet.}

\end{figure}

We compare our methods with methods including PSR~\cite{DBLP:journals/tog/KazhdanH13}, PSG~\cite{DBLP:conf/cvpr/FanSG17}, R2N2~\cite{DBLP:conf/eccv/ChoyXGCS16}, Atlas~\cite{groueix2018papier}, COcc~\cite{DBLP:conf/eccv/PengNMP020}, SAP~\cite{Peng2021SAP}, OCNN~\cite{wang2020deep}, IMLS~\cite{Liu2021MLS} and POCO~\cite{Boulch_2022_CVPR}. The numerical comparison in Tab.~\ref{table:shapenet3} demonstrates our state-of-the-art surface reconstruction accuracy over 13 classes. Although we do not require the ground truth supervision, our method outperforms the supervised methods such as SAP~\cite{Peng2021SAP}, COcc~\cite{DBLP:conf/eccv/PengNMP020} and IMLS~\cite{Liu2021MLS}. We further demonstrate our superiority in the reconstruction of complex geometry in the visual comparison in Fig.~\ref{fig:ShapeNet}. More numerical and visual comparisons can be found in our supplemental materials.

\begin{table*}[t]
\centering
\setlength{\tabcolsep}{0.5mm}

\resizebox{\linewidth}{!}{

    \begin{tabular}{c|cccccccccc|cccccccccc|cccccccccc}  %
     \toprule
     \multirow{1}*{-} & \multicolumn{10}{c|}{L1CD} & \multicolumn{10}{c|}{Normal Consistency} & \multicolumn{10}{c}{F-Score}\\
     \midrule
     -&PSR&PSG&R2N2&Atlas&COcc&SAP&OCNN&IMLS&POCO&\textbf{Ours}&PSR&PSG&R2N2&Atlas&COcc&SAP&OCNN&IMLS&POCO&\textbf{Ours}&PSR&PSG&R2N2&Atlas&COcc&SAP&OCNN&IMLS&POCO&\textbf{Ours}\\
     \midrule
     airplane&0.437&0.102&0.151&0.064&0.034&0.027&0.063&0.025&0.023&\textbf{0.022}&0.747&-&0.669&0.854&0.931&0.931&0.918&0.937&0.944&\textbf{0.960}&0.551&0.476&0.382&0.827&0.965&0.981&0.810&0.992&0.994&\textbf{0.995}\\
     bench&0.544&0.128&0.153&0.073&0.035&0.032&0.065&0.030&0.028&\textbf{0.025}&0.649&-&0.691&0.820&0.921&0.920&0.914&0.922&0.928&\textbf{0.935}&0.430&0.266&0.431&0.786&0.965&0.979&0.800&0.986&0.988&\textbf{0.993}\\
     cabinet&0.154&0.164&0.167&0.112&0.047&0.037&0.071&0.035&0.037&\textbf{0.034}&0.835&-&0.786&0.875&0.956&0.957&0.941&0.955&0.961&\textbf{0.975}&0.728&0.137&0.412&0.603&0.955&0.975&0.789&0.981&0.979&\textbf{0.996}\\
     car&0.180&0.132&0.197&0.099&0.075&0.045&0.077&0.040&0.041&\textbf{0.037}&0.783&-&0.719&0.827&0.893&0.897&0.867&0.882&0.894&\textbf{0.937}&0.729&0.211&0.348&0.642&0.849&0.928&0.747&0.952&0.946&\textbf{0.964}\\
     chair&0.369&0.168&0.181&0.114&0.046&0.036&0.066&0.035&0.033&\textbf{0.026}&0.715&-&0.673&0.829&0.943&0.952&0.941&0.950&0.956&\textbf{0.965}&0.473&0.152&0.393&0.629&0.939&0.979&0.799&0.982&0.985&\textbf{0.993}\\
     display&0.280&0.160&0.170&0.089&0.036&0.030&0.066&0.029&0.028&\textbf{0.022}&0.749&-&0.747&0.905&0.968&0.972&0.960&0.973&0.975&\textbf{0.981}&0.544&0.175&0.401&0.727&0.971&0.990&0.811&0.994&0.994&\textbf{0.998}\\
     lamp&0.278&0.207&0.243&0.137&0.059&0.047&0.067&0.031&0.033&\textbf{0.027}&0.765&-&0.598&0.759&0.900&0.921&0.911&0.922&0.929&\textbf{0.957}&0.586&0.204&0.333&0.562&0.892&0.959&0.800&0.979&0.975&\textbf{0.990}\\
     speaker&0.148&0.205&0.199&0.142&0.063&0.041&0.073&0.040&0.041&\textbf{0.033}&0.843&-&0.735&0.867&0.938&0.950&0.936&0.947&0.952&\textbf{0.977}&0.731&0.107&0.405&0.516&0.892&0.957&0.779&0.963&0.964&\textbf{0.977}\\
     rifle&0.409&0.091&0.167&0.051&0.028&0.023&0.062&0.021&0.019&\textbf{0.019}&0.788&-&0.700&0.837&0.929&0.937&0.932&0.943&\textbf{0.949}&0.938&0.590&0.615&0.381&0.877&0.980&0.990&0.826&0.996&0.998&\textbf{0.998}\\
     sofa&0.227&0.144&0.160&0.091&0.041&0.032&0.066&0.031&0.030&\textbf{0.027}&0.826&-&0.754&0.888&0.958&0.963&0.949&0.963&0.967&\textbf{0.978}&0.712&0.184&0.427&0.717&0.953&0.982&0.801&0.987&0.989&\textbf{0.992}\\
     table&0.393&0.166&0.177&0.102&0.038&0.033&0.066&0.032&0.031&\textbf{0.028}&0.706&-&0.734&0.867&0.959&0.962&0.946&0.962&0.966&\textbf{0.970}&0.442&0.158&0.404&0.692&0.967&0.986&0.801&0.987&0.991&\textbf{0.992}\\
     telephone&0.281&0.110&0.130&0.054&0.027&0.023&0.061&0.023&0.022&\textbf{0.017}&0.805&-&0.847&0.957&0.983&0.984&0.974&0.984&0.985&\textbf{0.987}&0.674&0.317&0.484&0.867&0.989&0.997&0.825&0.998&0.998&\textbf{0.999}\\
     vessele&0.181&0.130&0.169&0.078&0.043&0.030&0.064&0.027&0.025&\textbf{0.024}&0.820&-&0.641&0.837&0.918&0.930&0.922&0.932&0.940&\textbf{0.951}&0.771&0.363&0.394&0.7757&0.931&0.974&0.809&0.987&0.989&\textbf{0.997}\\
     \midrule
     mean&0.299&0.147&0.173&0.093&0.044&0.034&0.067&0.031&0.030&\textbf{0.026}&0.772&-&0.715&0.855&0.938&0.944&0.932&0.944&0.950&\textbf{0.962}&0.612&0.259&0.400&0.708&0.942&0.975&0.800&0.983&0.984&\textbf{0.991}\\
     \bottomrule
\end{tabular}}
\caption{L1CD, NC and F-Score comparison under ShapeNet.}
   \label{table:shapenet3}
\end{table*}%

\noindent\textbf{FAMOUS and ABC. }We further evaluate our method using the test set in FAMOUS and ABC dataset provided by P2S~\cite{ErlerEtAl:Points2Surf:ECCV:2020}. The clean point cloud is corrupted with noise at different levels. We follow NeuralPull~\cite{Zhizhong2021icml} to report L2 Chamfer Distance (L2CD). Different from previous experiments, we only leverage single $N=1$ noisy point clouds to train our method with a batch size of $B=1000$.

\begin{figure}[tb]

  \centering
   \includegraphics[width=\linewidth]{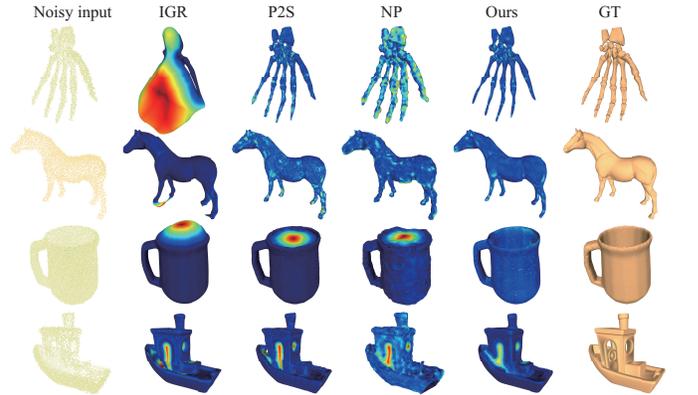}
  
\caption{\label{fig:famous}Visual comparison in surface reconstruction under FAMOUS. Point to surface error at each vertex is shown in color.}

\end{figure}

We compare our methods with methods including DSDF~\cite{Park_2019_CVPR}, Atlas~\cite{groueix2018papier}, PSR~\cite{DBLP:journals/tog/KazhdanH13}, P2S~\cite{ErlerEtAl:Points2Surf:ECCV:2020}, NP~\cite{Zhizhong2021icml}, IMLS~\cite{Liu2021MLS}, PCP~\cite{DBLP:conf/cvpr/MaLZH22}, POCO~\cite{Boulch_2022_CVPR}, and OnSF~\cite{DBLP:conf/cvpr/MaLH22}. 
\js{Note that the implementation of Chamfer Distance (CD) used in P2S~\cite{ErlerEtAl:Points2Surf:ECCV:2020} and occupancy network~\cite{MeschederNetworks} are different. For a comprehensive and fair comparison with the baselines evaluated under different settings, we report the performance of our method in both P2S-CD setting and the L2CD setting used in occupancy network, and make comparisons under both settings to demonstrate the effectiveness of our method.} The comparison in Tab.~\ref{table:NOX3noise} demonstrates that our method can reveal more accurate surfaces from noisy point clouds even we do not have training set, ground truth supervision or even multiple noisy point clouds. The statistical reasoning on point clouds and geometric regularization produce more accurate surfaces as demonstrated by the error map comparison under FAMOUS in Fig.~\ref{fig:famous}.

\begin{table}[t]
\centering
\setlength{\tabcolsep}{1.2mm}

\resizebox{\linewidth}{!}{

    \begin{tabular}{c|cccccc|ccccc}  %
     \toprule
     \multirow{1}*{-} & \multicolumn{6}{c|}{P2S-CD} & \multicolumn{5}{c}{L2CD}\\
     \midrule
     Dataset&DSDF&Atlas&PSR&P2S&POCO&Ours&NP& IMLS&OnSF&PCP&Ours\\
     \midrule
       ABC var& 12.51 & 4.04 & 3.29& 2.14&2.01& \textbf{1.87} & 0.72 & 0.57 &3.52&0.49&\textbf{0.113} \\ %
       ABC max& 11.34 & 4.47 & 3.89& 2.76&2.50 & \textbf{2.42} &1.24& -&4.30& 0.57&\textbf{0.139} \\ %
       \midrule
       F-med &9.89&4.54&1.80&1.51&1.50& \textbf{1.32} &0.28& 0.80&0.59&0.07&\textbf{0.033}\\
       F-max &13.17&4.14&3.41&2.52&2.75& \textbf{2.31} &0.31& - &3.64&0.30&\textbf{0.117}\\
     \bottomrule
\end{tabular}}
   \caption{\js{Comparison under ABC and Famous datasets with P2S-CD and L2CD settings.}}
   \label{table:NOX3noise}
\end{table}

\noindent\textbf{D-FAUST and SRB. }Finally, we evaluate our method under the real scanning dataset D-FAUST~\cite{dfaust:CVPR:2017} and SRB~\cite{berger2013benchmark}. We follow SAP~\cite{Peng2021SAP} to evaluate our result using L1CD, NC~\cite{MeschederNetworks}, and F-score~\cite{Tatarchenko_2019_CVPR} with a threshold of $1\%$ using the same set of shapes. We use single $N=1$ noisy point clouds to train our method with a batch size of $B=5000$.

\begin{figure}[tb]
  \centering
   \includegraphics[width=\linewidth]{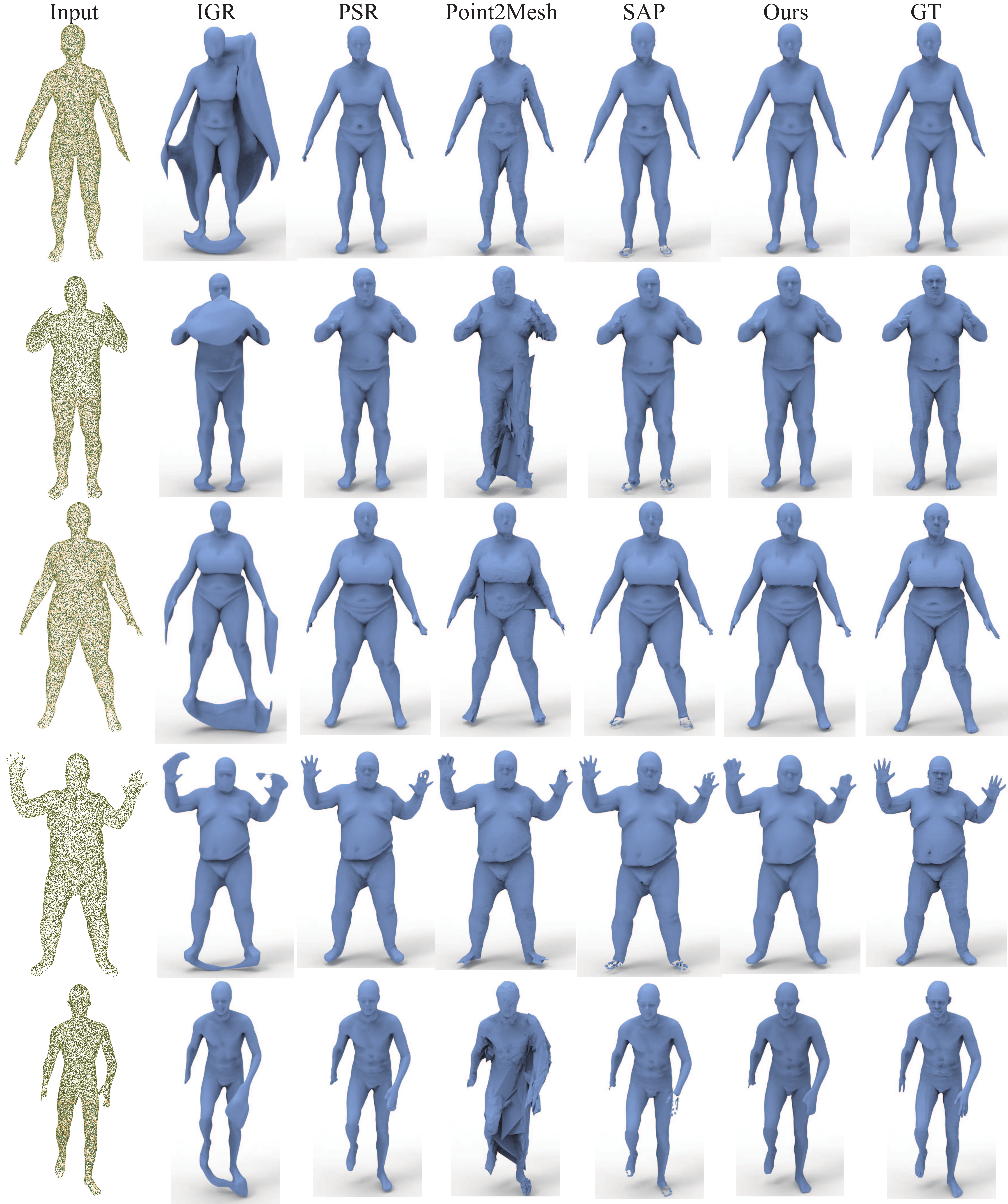}
  
\caption{\label{fig:DFAUST}Comparison in surface reconstruction under DFAUST.}

\end{figure}

\begin{figure}[tb]

  \centering
   \includegraphics[width=\linewidth]{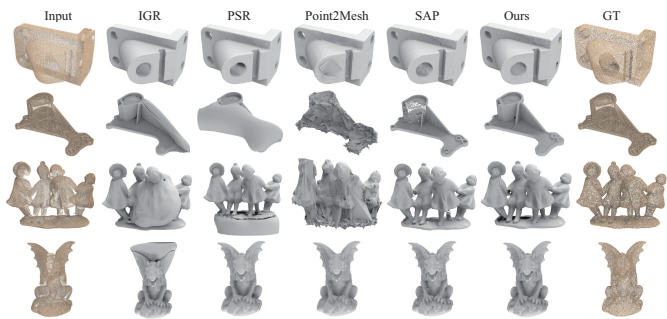}
  
\caption{\label{fig:SRB}Comparison in surface reconstruction under SRB.}

\end{figure}

\js{We compare our methods with the methods including IGR~\cite{DBLP:conf/icml/GroppYHAL20}, Point2Mesh~\cite{DBLP:journals/tog/HanockaMGC20}, PSR~\cite{DBLP:journals/tog/KazhdanH13}, NDF \cite{chibane2020neural}, NP \cite{ma2020neuralpull}, MLOD~\cite{mercier2022moving}, Neural-IMLS~\cite{wang2021neural}, SAP~\cite{Peng2021SAP} and DiGS \cite{ben2021digs}. We report numerical comparisons in Tab.~\ref{table:dfaust} and Tab.~\ref{table:srb}. Although we only do statistical reasoning on a single noisy point cloud and do not require point normals, our method still handles the noise in real scanning well, which achieves much smoother and more accurate structure.} The comparison in Fig.~\ref{fig:DFAUST} and Fig.~\ref{fig:SRB} shows that our method can produce more accurate surfaces without missing parts on both rigid and non-rigid shapes.

\begin{table}[t]
\centering
\resizebox{0.92\linewidth}{!}{
    \begin{tabular}{c|ccccc}  %
     \toprule
     Metrics&IGR&Point2Mesh&PSR&SAP&\textbf{Ours}\\
     \hline
     L1CD$\times 10$&0.235&0.071&0.044&0.043&\textbf{0.037}\\
     F-Score&0.805&0.855&0.966&0.966&\textbf{0.996}\\
     NC&0.911&0.905&0.965&0.959&\textbf{0.970}\\
     \toprule
\end{tabular}%
   }
   
   \caption{Comparison in surface reconstruction under D-FAUST.}
   
   \label{table:dfaust}
\end{table}

\begin{table}[t]
\setlength{\tabcolsep}{1.5mm}

\centering
\resizebox{0.9\linewidth}{!}{
    \begin{tabular}{c|ccccccc}  %
     \toprule
     Metrics&IGR&Point2Mesh&PSR&NDF&NP\\
     \hline
     L1CD$\times 10$&0.178&0.116&0.232&0.185&0.106\\
     F-Score&0.755&0.648&0.735&0.722&0.797\\
     \bottomrule
     \toprule
     Metrics&MLOD&Neural-IMLS&SAP&DiGS&\textbf{Ours}\\
     \hline
     L1CD$\times 10$&0.104&0.075& 0.076& 0.069 &\textbf{0.067}\\
     F-Score&0.765&0.822 &0.830&\textbf{0.839}&{0.835}\\
     \bottomrule
\end{tabular}%
   }
   
   \caption{\js{Comparison in surface reconstruction under SRB.}}
   
   \label{table:srb}
\end{table}

\subsection{Surface Reconstruction for Scenes}
\label{6.4}

\begin{figure*}[tb]
  \centering
   \includegraphics[width=\linewidth]{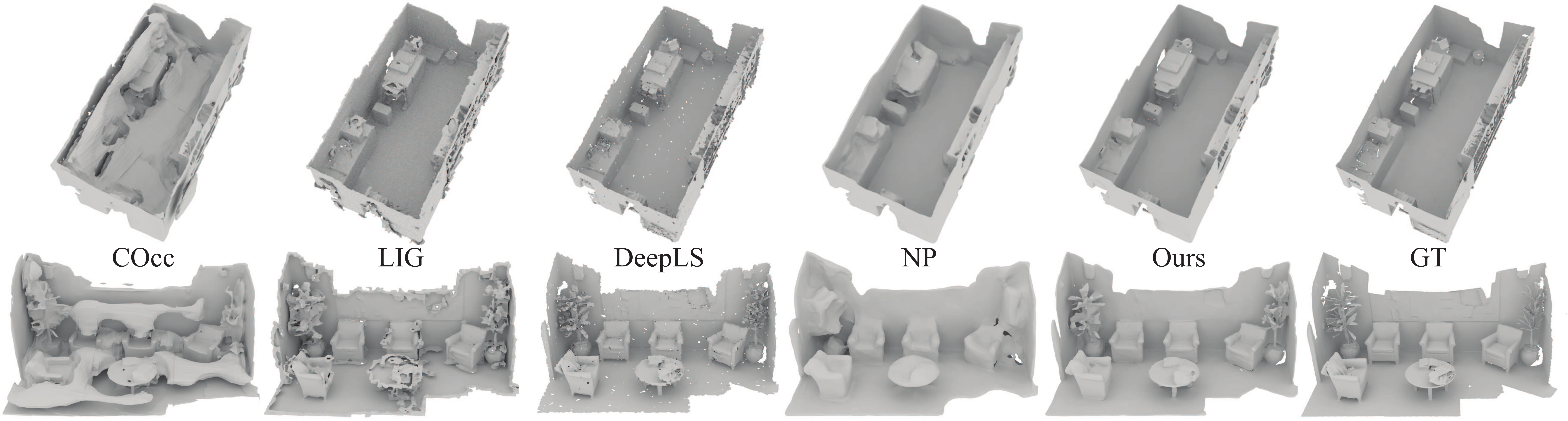}
  
\caption{\label{fig:Scenes}Visual comparison in surface reconstruction under 3D Scene dataset.}
\end{figure*}

\noindent\textbf{3D Scene. }We evaluate our method under real scene scan dataset~\cite{DBLP:journals/tog/ZhouK13}. We sample $1000$ points per $m^2$ from Lounge and Copyroom, and only leverage $N=1$ noisy point cloud to train our method with a batch size of $B=5000$. \js{We compare our method with SOTA data-driven based methods and overfitting based methods including LIG~\cite{jiang2020lig}, COcc~\cite{Peng2020ECCV}, NP~\cite{ma2020neuralpull}, DeepLS~\cite{DBLP:conf/eccv/ChabraLISSLN20}, POCO~\cite{pococvpr2022}, ALTO~\cite{wang2023alto}, GridFormer~\cite{li2024gridformer} and NKSR~\cite{huang2023neural}. We leverage the pretrained models of COcc, LIG, POCO, ALTO, GridFormer and NKSR, and retrain NP and DeepLS to produce their results with the same input. Numerical comparison in Tab.~\ref{table:t12} demonstrates that our method outperforms the state-of-the-art results. Fig.~\ref{fig:Scenes} further demonstrates that we can produce much smoother surfaces with more geometry details. For more analysis and qualitative comparisons under 3DScene dataset, please refer to the supplementary materials.}

\begin{table}[t]
\centering
\resizebox{\linewidth}{!}{
    \begin{tabular}{c|c|c|c||c|c|c}
     \hline

        &\multicolumn{3}{c||}{Lounge}&\multicolumn{3}{c}{Copyroom}\\
        \hline
       &L2CD&L1CD&NC&L2CD&L1CD&NC\\
     \hline
     COcc~\cite{DBLP:conf/eccv/PengNMP020}&9.540 &0.046 &0.894 &10.97 &0.045 &0.892\\
     LIG~\cite{jiang2020lig}&9.672 &0.056 &0.833 &3.61 &0.036 &0.810\\
     DeepLS~\cite{DBLP:conf/eccv/ChabraLISSLN20}&6.103&0.053&0.848&0.609&0.021&0.901\\
     NP~\cite{Zhizhong2021icml}&1.079 &0.019 &0.910 &5.795 &0.036 &0.862\\
     POCO~\cite{Boulch_2022_CVPR} & 1.122 & 0.024 & 0.912 & 1.468 & 0.025 & 0.883 \\
     ALTO~\cite{wang2023alto} & 1.716 & 0.033 & 0.904 & 1.095 & 0.028 & 0.839 \\
     GridFormer~\cite{li2024gridformer} & 1.345 & 0.028 & 0.901 & 3.485 & 0.033 & 0.864 \\
     NKSR~\cite{huang2023neural} & 0.617 & 0.019 & 0.917 & 0.527 & 0.017 & \textbf{0.908} \\
     
     \hline
     Ours&\textbf{0.602}&\textbf{0.016}&\textbf{0.923}&\textbf{0.442}&\textbf{0.016}&{0.903}\\
     \hline
   \end{tabular}}
   
   \caption{\js{Surface reconstruction under 3D Scene dataset. L2-CD$\times 10^3$. The unit of error is mm.}}
   
   \label{table:t12}
\end{table}

\begin{figure}[tb]
  \centering
   \includegraphics[width=1\linewidth]{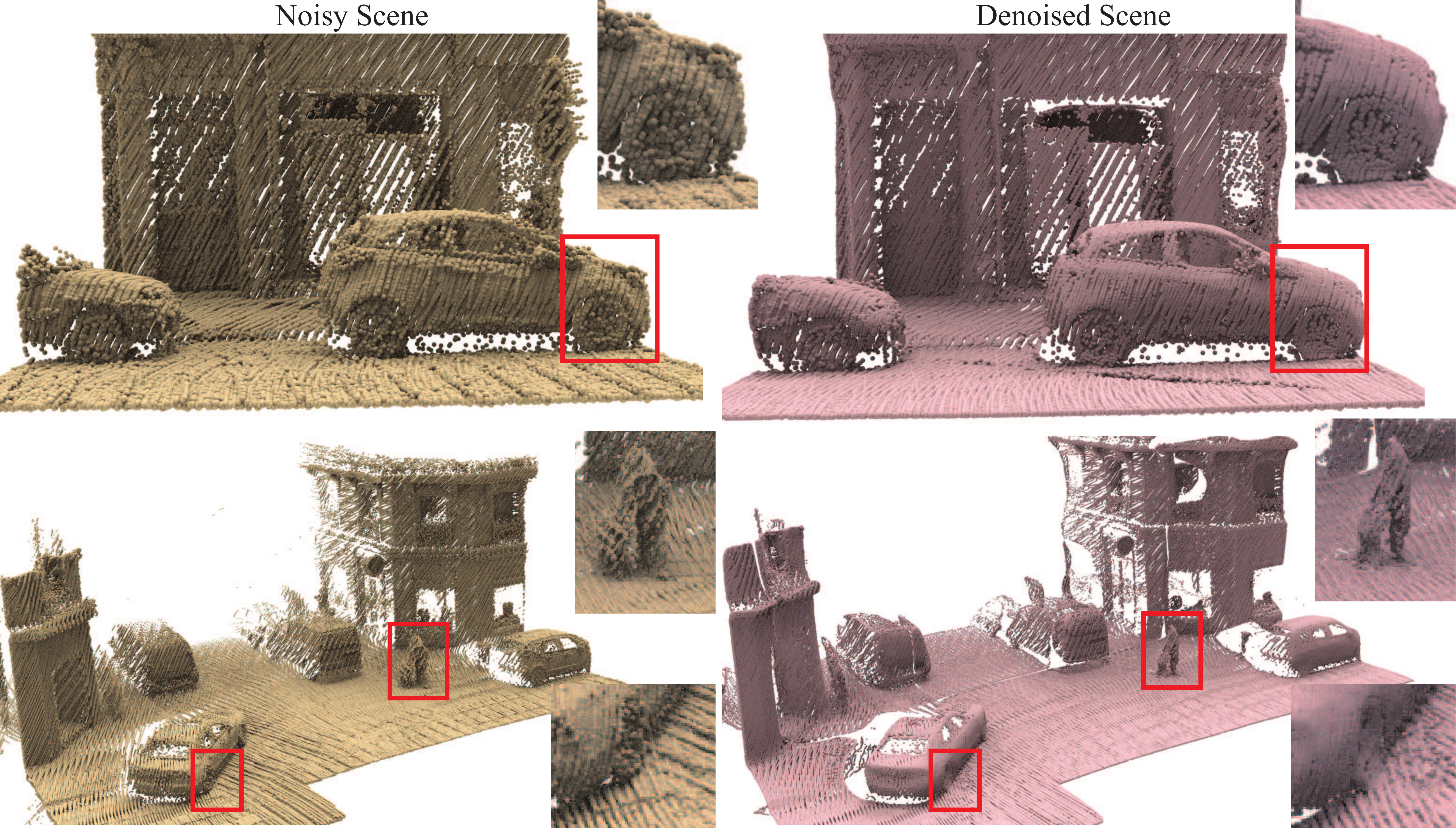}
  
\caption{\label{fig:ParisZoom}Demonstration of denoising on real scans.}

\end{figure}

\begin{figure*}[t]
  \centering
   \includegraphics[width=0.9\linewidth]{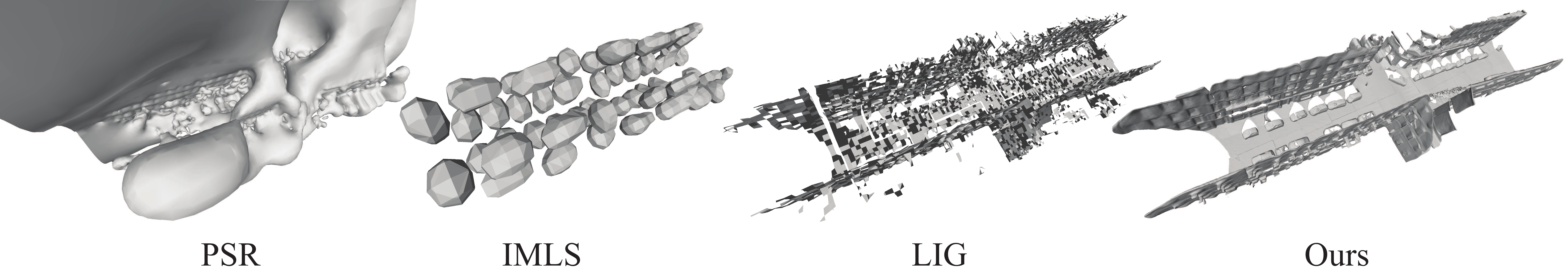}
  
\caption{\label{fig:Paris}Comparison in surface reconstruction from real scans.}

\end{figure*}

\noindent\textbf{Paris-rue-Madame. }We further evaluate our method under another real scene scan dataset~\cite{DBLP:conf/icpram/SernaMGD14}. We only use $N=1$ noisy point cloud with a batch size of $B=5000$. We split the $10M$ points into $50$ chunks each of which is used to learn a SDF. Similarly, we use each chunk to evaluate IMLS~\cite{Liu2021MLS} and LIG~\cite{jiang2020lig} with their pretrained models. Our superior performance over the latest methods in large scale surface reconstruction is demonstrated in Fig.~\ref{fig:Paris}. Our denoised point clouds in a smaller scene are detailed in Fig.~\ref{fig:ParisZoom}.

\begin{figure}[tb]
  \centering
   \includegraphics[width=\linewidth]{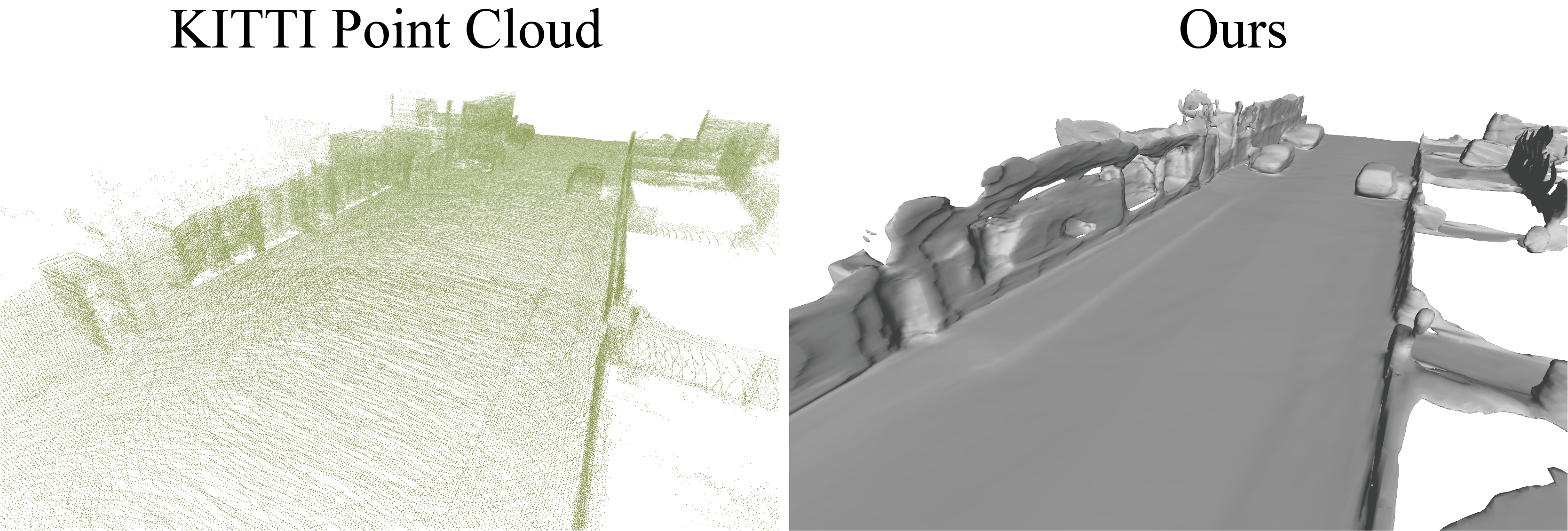}
\caption{\label{fig:KITTI}Reconstruction on a real scan from KITTI.}
\end{figure}

\noindent\textbf{KITTI.} Additionally, we report our reconstruction on a road from KITTI in Fig.~\ref{fig:KITTI}. Our method can also reconstruct plausible and smooth surfaces from a single real scan containing sparse and noisy points captured by vehicular radars.

\begin{table}[t]
\centering
\resizebox{\linewidth}{!}{
    \begin{tabular}{c|c|c||c|c||c}
     \hline

        &\multicolumn{2}{c||}{Lounge}&\multicolumn{2}{c||}{Copyroom}&\multicolumn{1}{c}{Training}\\
        \cline{2-5}
       &L2CD&L1CD&L2CD&L1CD&time\\
     \hline
     DeepLS~\cite{DBLP:conf/eccv/ChabraLISSLN20}&6.103&0.053&0.609&0.021&-\\
     NP~\cite{Zhizhong2021icml}&1.079 &0.019 &5.795 &0.036 & 1150 s\\
     SAP~\cite{Peng2021SAP}&1.801&0.025&1.992&0.027& 288 s\\
     
     \hline
     Ours&0.602&\textbf{0.016}&\textbf{0.442}&0.016& 913 s\\
     Ours (+Fast)&\textbf{0.496}&0.020&0.457&\textbf{0.015}&\textbf{63 s}\\
     \hline
   \end{tabular}}
   
   \caption{Surface reconstruction under 3D Scene dataset. L2-CD$\times 10^3$. The unit of error is mm.}
   
   \label{table:fast1}
\end{table}

\begin{table}[t]
\centering
\resizebox{\linewidth}{!}{
    \begin{tabular}{c|c|c||c|c||c}
     \hline

        &\multicolumn{2}{c||}{10K(Sparse)}&\multicolumn{2}{c||}{50K(Dense)}&\multicolumn{1}{c}{Training}\\
        \cline{2-5}
       &CD&P2M&CD&P2M&time\\
     \hline
     Ours&\textbf{4.221}&1.847&\textbf{1.654}&\textbf{0.972}& 1308 s\\
     Ours (+Fast)&4.537&\textbf{1.723}&1.889&1.273& \textbf{102 s}\\
     \hline
   \end{tabular}}
   
   \caption{Point cloud denoising under PUNet dataset. L2CD$\times 10^4$ and P2M $\times 10^4$.}
   
   \label{table:fast2}
\end{table}

\subsection{\md{Fast Learning Reconstruction and Denoising}}
\label{6.5}

We evaluate our fast learning framework in point cloud denoising and scene reconstruction tasks. The metrics and experiment settings keep the same as Sec. \ref{6.1} and Sec. \ref{6.4}. All timing results are reported on one single RTX3090 GPU.

\noindent\textbf{Fast Learning Reconstruction.} We conduct experiments under 3D Scene dataset for evaluating the surface reconstruction ability of our fast learning framework. Numerical comparisons in Tab. \ref{table:fast2} show that our fast learning framework can reconstruct complex scenes in about one minute and achieves comparable performance to our origin MLP-based approach for about 15 minutes. Our method also outperforms the state-of-the-art work on fast reconstruction, i.e. SAP \cite{Peng2021SAP}, in terms of both speed and qualities as shown in Tab. \ref{table:fast2}. The visualizations under different optimization time with or without our fast learning framework are shown in Fig. \ref{fig:ngp_results}, where we can achieve a visual appearing reconstruction in 6 seconds with the multi-resolution hash encodings. 

\noindent\textbf{Fast Learning Denoising.} We further evaluate our fast learning framework for point cloud denoising under PU dataset. Same as Sec. \ref{6.1}, we report our results under 20 shapes in the test set of PU with different densities of point clouds, i.e., 10K and 50K. We evaluate the results under large Gaussian noise with a standard deviation of 3\%. The comparisons in Tab. \ref{table:fast1} show that our fast learning framework (+Fast) achieves comparable performance to our origin MLP-based approach, yet is more than 10 times faster, which demonstrates the effectiveness and efficiency of our proposed fast learning framework.

\subsection{\md{Multi-view Reconstruction}}
\label{6.6}

\md{Neural surface reconstruction from multi-view images \cite{Oechsle2021ICCV,yariv2020multiview,yariv2021volume,neuslingjie,cai2023neuda,liu2023neudf,zhou2023levelset,li2024LDI,long2023neuraludf,li2023neuralangelo,rosu2023permutosdf,xiang2021snowflake,wen2022pmp,zhang2023fast} has been shown to be powerful for recovering dense 3D surfaces via image-based neural rendering. We demonstrate that our method can also improve the performance of multi-view reconstruction by introducing an SDF through noise to noise mapping on point clouds from SfM \cite{schoenberger2016sfm} as a geometry prior.}

\noindent\textbf{Dataset and Metrics.} To evaluate the effectiveness of our proposed schema to learn SDF priors for multi-view reconstruction, we conduct experiments under the DTU dataset \cite{jensen2014large} to reconstruct surfaces from multi-view images. Following previous works \cite{Oechsle2021ICCV,yariv2020multiview,yariv2021volume,neuslingjie}, we report the results under the widely used 15 scenes. Each of these scenes captures 49 to 64 images  around a single object with background. For quantitative evaluations under the DTU dataset, we first clean the reconstructed meshes using the provided masks following previous works \cite{Oechsle2021ICCV,yariv2020multiview,yariv2021volume, neuslingjie}, and then leverage L1 Chamfer distance as the metric to evaluate the errors of the randomly sampled points on the reconstructed surfaces compare to the ground truth point clouds. 

\begin{figure*}[tb]
  \centering
   \includegraphics[width=\linewidth]{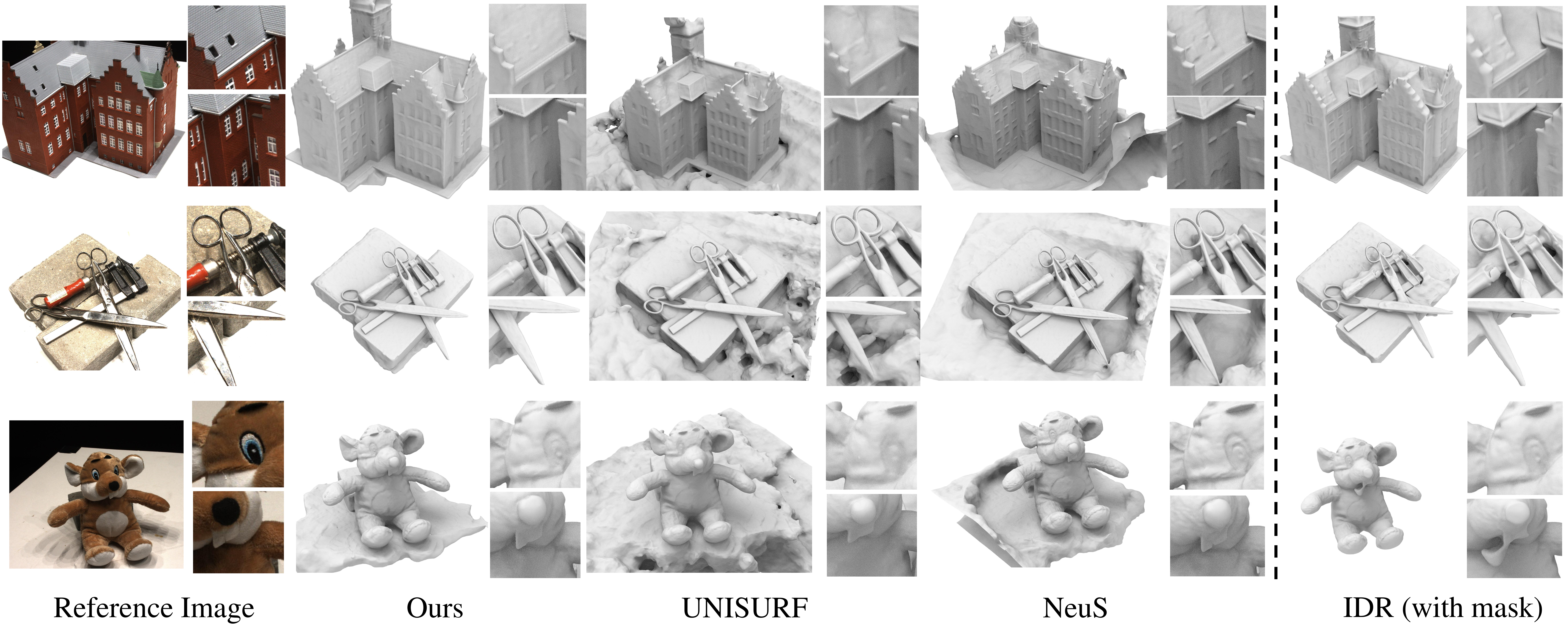}
\caption{\label{fig:dtu} Qualitative comparisons on DTU dataset. IDR requires masks as additional input.}
\end{figure*}

\begin{figure}[tb]
  \centering
   \includegraphics[width=\linewidth]{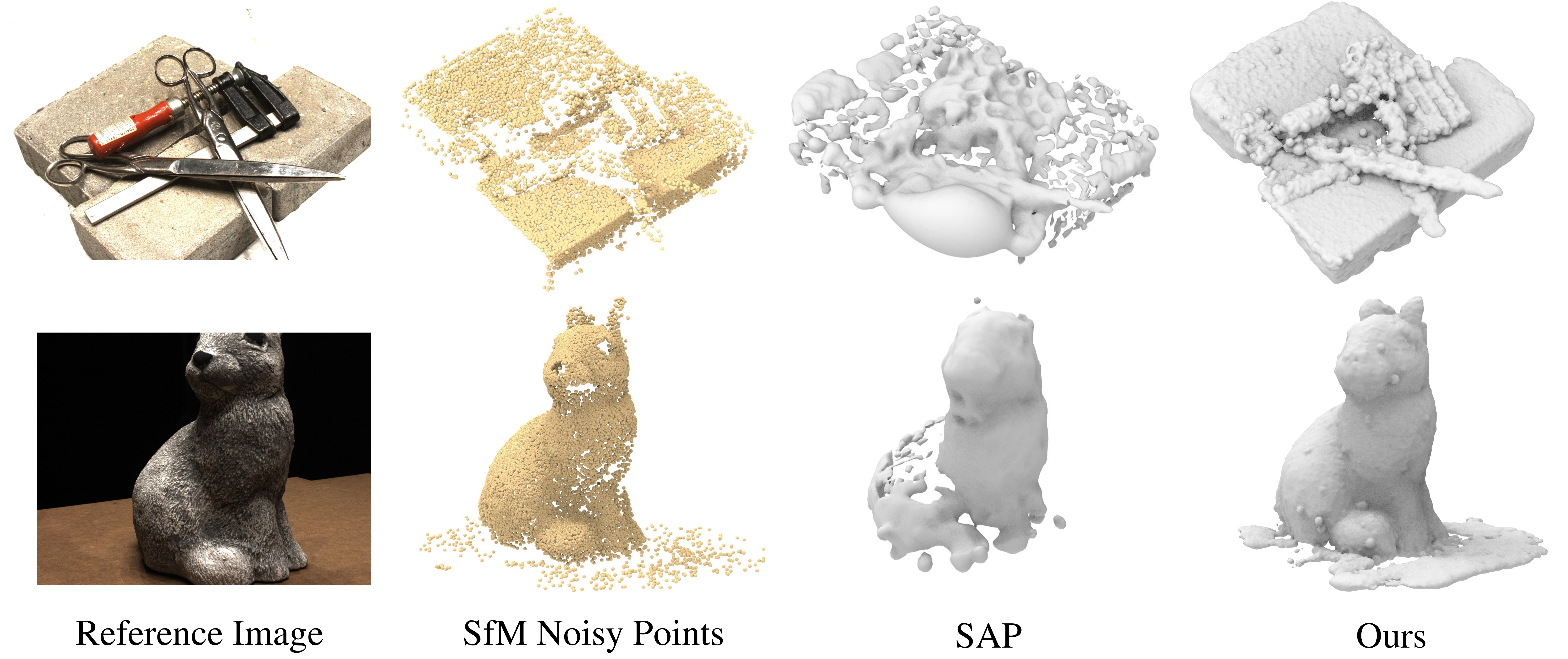}
\caption{\label{fig:dtu_prior} Comparison in the learned SDF priors from noisy SfM point clouds. }
\end{figure}

\noindent\textbf{Comparison.} We compare our method with the traditional method Colmap \cite{schoenberger2016sfm} and the state-of-the-art learning-based methods NeuS \cite{neuslingjie}, UNISURF \cite{Oechsle2021ICCV}, VolSDF \cite{yariv2021volume}, Voxurf \cite{wu2022voxurf}, MonoSDF \cite{Yu2022MonoSDF} and HF-NeuS \cite{yiqunhfSDF}. The numerical comparisons are reported in Tab. \ref{tab:dtu}, where our method achieves better performance than the state-of-the-art methods. It is worth noticing that our method, which is built upon NeuS \cite{neuslingjie}, brings large performance gain than the baseline NeuS (0.87$\rightarrow$0.72). This result demonstrates the effectiveness of our proposed schema which learns a clean field prior from noisy points as priors to serve as a guidance for multi-view learning.  We also produce better reconstruction results than MonoSDF \cite{Yu2022MonoSDF}, which introduces both depth and normal priors from pretrained omnidata models learned using large-scale datasets as additional supervisions. The results demonstrate that our SDF priors perform better compared with the monocular priors achieve by large-scale data.   

We further show visual comparisons with the state-of-the-arts before cleaning with GT masks in Fig. \ref{fig:dtu}, where our method not only produces more accurate reconstructions, but also largely remove the artifacts in empty space. The reason is that our SDF priors also serves as a regularization constraint to guide the learning of SDFs at the empty space. More specifically, when getting supervised by our SDF priors, noisy points in the empty space will be removed. Note that the performance gain in removing artifacts is not reflected in the numerical results since the artifacts in empty space has been cleaned using the GT masks following the evaluation protocol.

We then provide a visual comparison on learning SDF priors from SfM noisy point clouds by visualizing the reconstructions of the learned SDF priors in Fig. \ref{fig:dtu_prior}. The results show that our methods can learn a clean and complete SDF prior from the noisy input, where the other methods (e.g. SAP \cite{Peng2021SAP}) lead to failures using noisy point clouds. The clean SDF prior helps NeuS \cite{neuslingjie} to learn the exact geometry information provided by SfM and also serve as a artifact-free field initialization which regularizes the optimization of multi-view reconstruction.

\begin{figure*}[t]
  \centering

   \includegraphics[width=\linewidth]{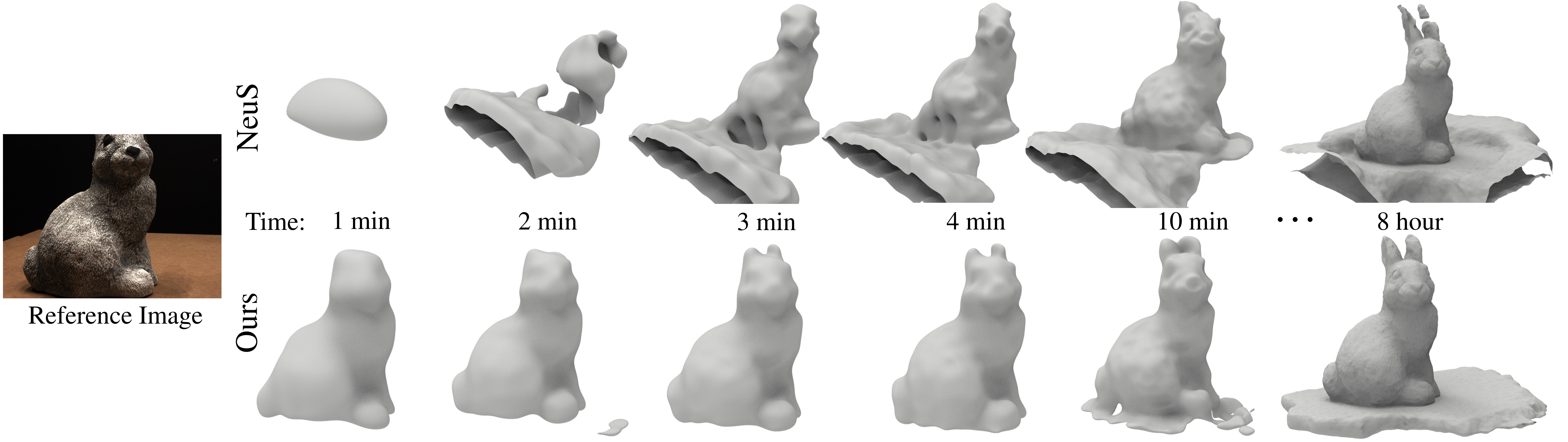}

\caption{\label{fig:time_dtu} Visual comparisons on multi-view reconstructions under different optimization time. The timing results do not include the time for Marching Cubes, and are reported under one single RTX3090 GPU.}

\end{figure*}

To further demonstrate the efficiency of our proposed method, we make a comparison with NeuS \cite{neuslingjie} on the reconstructions at different optimization time in Fig. \ref{fig:time_dtu}. We exclude the time to get points and only show the time for optimization. The result shows that our method achieves much faster convergence than NeuS, where we can obtain a coarse reconstruction in one minute with the clean field prior. And the field prior also helps to learn accurate geometries, i.e., the complete rabbit ears and the clean and closed ground. 
There are some recent studies \cite{wang2022neus2, zhao2022human} focus on accelerating NeuS by improving the framework of NeuS. We justify that our method have no intersection with these methods since we do not have special designs on the framework, but using SDF priors which can also be integrated into these methods to further improve the reconstruction qualities and training efficiency.

\setlength\tabcolsep{6pt}
\begin{table*}[th!]
\centering
\resizebox{\linewidth}{!}{
\begin{tabular}{ c | c c c c c c c c c c c c c c c | c }
\toprule
ScanID & 24 & 37 & 40 & 55 & 63 & 65 & 69 & 83 & 97 & 105 & 106 & 110 & 114 & 118 & 122 & Mean \\

\midrule
$\text{colmap}$ \cite{schoenberger2016sfm} & 0.81 & 2.05 & 0.73 & 1.22 & 1.79 & 1.58 & 1.02 & 3.05 & 1.4 & 2.05 & 1.00 & 1.32 & 0.49 & 0.78 & 1.17 & 1.36 \\
IDR$*$ \cite{yariv2020multiview} & 1.63 & 1.87 & 0.63 &0.48 &1.04 &0.79 & 0.77 &1.33 & 1.16 & 0.76 &0.67 & 0.90 & 0.42 & 0.51 & 0.53 & 0.90 \\
UNISURF \cite{Oechsle2021ICCV} & 1.32 & 1.36 & 1.72 & 0.44 & 1.35 & 0.79 & 0.80 & 1.49 & 1.37 & 0.89 & 0.59 & 1.47 & 0.46 & 0.59 & 0.62 & 1.02 \\
VolSDF \cite{yariv2021volume} & 1.14 & 1.26 & 0.81 & 0.49 & 1.25 & 0.70 & 0.72 & 1.29 & 1.18 & 0.70 & 0.66 & 1.08 & 0.42 & 0.61 & 0.55 & 0.86 \\
Voxurf \cite{wu2022voxurf} & 0.71 & 0.78 & 0.43 & 0.35 & 1.03 & 0.76 & 0.74 & 1.49 & 1.04 & 0.74 & 0.51 & 1.12 & 0.41 & 0.55 & 0.45 & 0.74 \\
NeuS \cite{neuslingjie} & 1.37 & 1.21 & 0.73 & 0.40 & 1.20 & 0.70 & 0.72 & 1.01 & 1.16 & 0.82 & 0.66 & 1.69 & 0.39 & 0.49 & 0.51 & 0.87 \\
MonoSDF(MLP) \cite{Yu2022MonoSDF} & 0.83 & 1.61& 0.65& 0.47& 0.92& 0.87& 0.87& 1.30& 1.25& 0.68& 0.65& 0.96& 0.41& 0.62& 0.58& 0.84 \\
MonoSDF(Grid) \cite{Yu2022MonoSDF} & 0.66 &0.88 &0.43 &0.40 &0.87 &0.78 &0.81 &1.23 &1.18 &0.66 &0.66 &0.96 &0.41 &0.57 &0.51 &0.73 \\
HF-NeuS \cite{yiqunhfSDF} & 0.76 & 1.32 & 0.70 & 0.39 & 1.06 & 0.63 & 0.63 & 1.15 & 1.12 & 0.80 & 0.52 & 1.22 & 0.33 & 0.49 & 0.50 & 0.77 \\
\midrule

Ours & 0.72 & 0.79 & 0.58 & 0.39 & 0.96 & 0.64 & 0.62 & 1.41 & 0.89 & 0.82 & 0.53 & 1.13 & 0.34 & 0.51 & 0.48 & \textbf{0.72} \\
\bottomrule
\end{tabular}}
\caption{Quantitative Comparisons on DTU. We compare the proposed method to the state-of-the-art baselines, \emph{i.e.}, NeuS and HF-NeuS, and other methods using their released codes following their best configurations. IDR$*$ means that the ground truth masks are required for IDR, where our method and other baselines do not use masks for training. }

\label{tab:dtu}
\end{table*}

\begin{table}[tb]
\centering
\resizebox{\linewidth}{!}{
    \begin{tabular}{c|c|c|c|c|c|c}  %
     \hline
          $B$ & 100 & 250 & 1000 & 2000 & 5000 & 10000\\   %
     \hline
       L2CD$\times10^4$&12.398 &\textbf{4.221}&4.578&5.628&5.998&6.217\\ %
       P2M$\times10^4$&5.482 &\textbf{1.847}&1.901&2.112&2.221&2.342\\
       \hline
   \end{tabular}}
   
   \caption{Effect of batch size $B$ under PU.}
   
   \label{table:batchsize}
\end{table}

\subsection{Ablation Studies}
\label{6.7}

\subsubsection{Training Settings}
We conduct ablation studies under the test set of PU. We first explore the effect of batch size $B$, training iterations, and the number $N$ of noisy point clouds in point cloud denoising. Tab.~\ref{table:batchsize} indicates that more points in each batch will slow down the convergence. Tab.~\ref{table:iters} demonstrates that more training iterations help perform statistical reasoning better to remove noise. Tab.~\ref{table:effectN} indicates that more corrupted observations are the key to increase the performance of statistical reasoning although one corrupted observation is also fine to perform statistical reasoning well.

\begin{table}[h]
\centering
\resizebox{0.92\linewidth}{!}{
    \begin{tabular}{c|c|c|c|c}  %
     \hline
          Iterations $\times 10^4$& 40 & 60 & 80 & 100 \\   %
     \hline
       L2CD$\times10^4$&4.887&4.364&\textbf{4.221}&4.224\\ %
       P2M$\times10^4$&2.032&1.885&\textbf{1.847}&1.849\\
       \hline
   \end{tabular}}
   
   \caption{Number of training iterations under PU.}
   
   \label{table:iters}
\end{table}

\begin{table}[h]
\centering
\setlength\tabcolsep{1mm}

\resizebox{\linewidth}{!}{
    \begin{tabular}{c|c|c|c|c|c|c|c}  %
     \hline
          $N$& 1& 2 & 10 & 20 & 50 &100 &200\\   %
     \hline
       L2CD$\times10^4$&4.976&4.898&4.665&4.558&4.432&4.224 &\textbf{4.221}\\ %
       P2M$\times10^4$&2.132&2.079&1.997&1.996&1.899&\textbf{1.847} &\textbf{1.847}\\
       \hline
   \end{tabular}}
   
   \caption{Effect of $N$ under PU.}
   
   \label{table:effectN}
\end{table}

\subsubsection{Loss Designs}
We further highlight the effect of EMD as the distance metric $L$ and geometric consistency regularization $R$ in denoising and surface reconstruction in Tab.~\ref{table:loss}. The comparison shows that we can not perform statistical reasoning on point clouds using CD, and EMD can only reveal the surface in statistical reasoning for denoising but not learn meaningful signed distance fields without $R$. Moreover, we found the $\lambda$ weighting $R$ slightly affects our performance.

\begin{table}[!]
\setlength\tabcolsep{1mm}

\centering
\resizebox{\linewidth}{!}{
    \begin{tabular}{c|c|c|c|c|c}  %
     \hline
          &CD&EMD,$\lambda=0$&EMD,$\lambda=0.05$&EMD,$\lambda=0.1$&EMD,$\lambda=0.2$\\
     \hline
       Denoise&73.786&\textbf{4.221}&4.245&4.252&4.832\\ %
       Reconstruction&81.573&80.917&5.721&\textbf{4.277}&4.993\\
       \hline
   \end{tabular}}
   
   \caption{Effect of CD and EMD as the distance metric $L$ and geometry consistency regularization $R$ under PU. L2CD$\times10^4$.}
   \label{table:loss}
\end{table}

\subsubsection{Number of Noisy Point Clouds}
We report additional ablation studies to explore the effect of the number of noisy point clouds in all the three tasks including point cloud denoising, point cloud upsampling, and surface reconstruction under the PU test set in Tab. \ref{table:point_num}. We can see we achieve the best performance with 200 noisy point clouds in all tasks, and the improvement over 100 point clouds is small. So we used 200 to report our results with multiple noisy point clouds in our paper.

\makeatletter
\newcommand\figcaption{\def\@captype{figure}\caption}
\newcommand\tabcaption{\def\@captype{table}\caption}
\makeatother

\begin{figure*}[tb]
\setlength{\tabcolsep}{1mm}

	\centering
		\begin{minipage}{0.49\textwidth}
			\centering
        \resizebox{\linewidth}{!}{
    \begin{tabular}{c|c|c|c|c|c|c|c|c}
     \hline
          $L$ & Metric & 1 & 2 & 10 & 20 & 50 & 100&200\\
     \hline
       \multirow{2}{*}{Denoise}&L2CD$\times10^4$&4.976&4.898&4.665&4.558&4.432	&4.224&\textbf{4.221}\\
       &P2M$\times10^4$&2.132&2.079&1.997&1.996&1.899&\textbf{1.847}&\textbf{1.847}\\
      \hline
       \multirow{2}{*}{Reconstruction}&L2CD$\times10^4$&5.102&4.995&4.795&4.599&4.456&4.369 &\textbf{4.355}\\
       &P2M$\times10^4$&2.423&2.217&2.007&2.001&1.978&1.886&\textbf{1.877}\\
      \hline
       \multirow{2}{*}{UpSampling}&L2CD$\times10^4$&4.988&4.886&4.687&4.574&4.461	&4.328&\textbf{4.272}\\
       &P2M$\times10^4$&2.152&2.082&2.001&1.997&1.977&1.919
 &\textbf{1.897}\\
      \hline
   \end{tabular}}
			\tabcaption{Effect of noisy point cloud numbers $L$ under PU.}
			\label{table:point_num}
		\end{minipage}
		\begin{minipage}{0.49\textwidth}
			\centering
        \resizebox{\linewidth}{!}{
    \begin{tabular}{c|c|c|c|c|c|c|c|c}
     \hline
          $D$& Metric & 1K & 2K & 5K & 10K & 20K & 50K &100K \\
     \hline
       \multirow{2}{*}{Denoise} & L2CD$\times10^4$&5.168&5.098&4.850&4.221&2.312&1.654 &\textbf{1.543}\\
       &P2M$\times10^4$&2.223&2.179&2.097&1.847&1.229&0.972 &\textbf{0.959}\\
       \hline
       \multirow{2}{*}{Reconstruction} & L2CD$\times10^4$&5.445&5.283&4.981&4.355&2.388&1.691 &\textbf{1.579}\\
       &P2M$\times10^4$&2.330&2.212&2.159&1.877&1.292&0.998 &\textbf{0.982}\\
       \hline
       \multirow{2}{*}{UpSampling} & L2CD$\times10^4$&5.281&5.187&4.984&4.272&2.392&1.682 &\textbf{1.561}\\
       &P2M$\times10^4$&2.398&2.212&2.167&1.897&1.289&0.997 &\textbf{0.973}\\
       \hline
   \end{tabular}}
			\tabcaption{Effect of Density $D$ of point Cloud under PU.}
			\label{table:density}
		\end{minipage}

\end{figure*}

\subsubsection{Point Density}
We report the effect of point density in all the three tasks including point cloud denoising, point cloud upsampling, and surface reconstruction under the PU test set in Tab. \ref{table:density}. We learn an SDF from a single noisy point cloud. With more noises, our method can achieve better performance in all the three tasks.

\subsubsection{One Observation vs. Multiple Observations.}
Since our method can learn from multiple observations and single observation, we investigate the effect of learning from these two training settings. Here, we combine multiple noisy observations into one noisy observation by concatenation, where we keep the total number of points the same. Table.~\ref{table:mixing} indicates that there is almost no performance difference with these two training settings.

\begin{table}[h]
\centering
\caption{Effect of mixing multiple noise point clouds under PU.}
\resizebox{\linewidth}{!}{
    \begin{tabular}{c|c|c|c}
     \hline
          Strategy& Metric & Mixing & W/O Mixing  \\
     \hline
       \multirow{2}{*}{Denoise} & L2CD$\times10^4$&4.244&\textbf{4.221}\\
       &P2M$\times10^4$&1.851&\textbf{1.847}\\
       \hline
       \multirow{2}{*}{Reconstruction} & L2CD$\times10^4$&\textbf{4.315}&4.355\\
       &P2M$\times10^4$&\textbf{1.831}&1.877\\
       \hline
       \multirow{2}{*}{UpSampling} & L2CD$\times10^4$&4.299&\textbf{4.272}\\
       &P2M$\times10^4$&\textbf{1.897}&\textbf{1.897}\\
       \hline
   \end{tabular}}
   \label{table:mixing}
\end{table}

\subsubsection{Point Noise Level}
\begin{figure}[h]
  \centering
   \includegraphics[width=\linewidth]{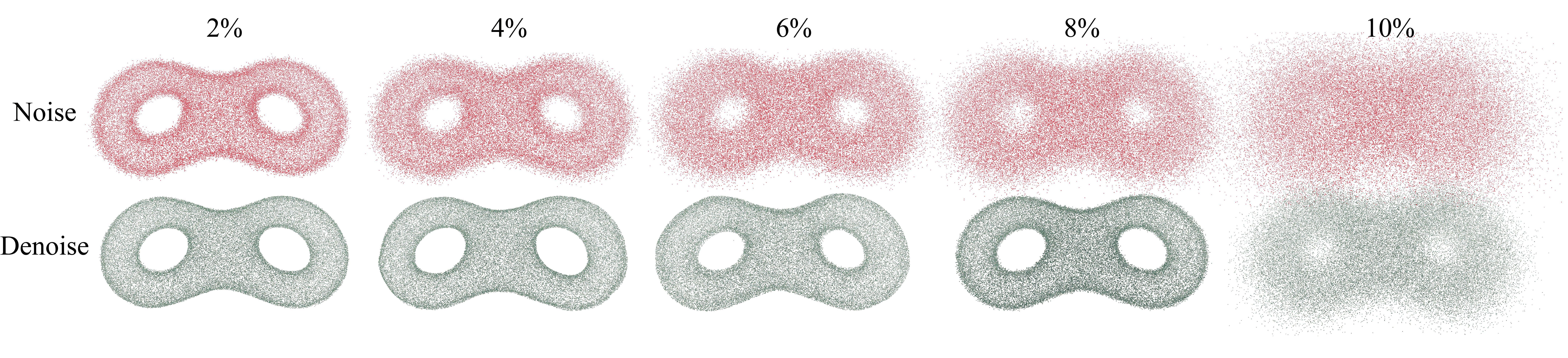}
\caption{\label{fig:noise_level}Point clouds denoising with large noises.}
\end{figure}

Additionally, we visualize our results with larger noises which we use to learn an SDF in point cloud denoising in Fig.~\ref{fig:noise_level}. We tried noises with different variances including $\{2\%,4\%,6\%,8\%,10\%\}$. We can see that our method can reveal accurate geometry with large noises. While our method may fail if the noises are too large to observe the structures, such as the variance of {10\%}. Note that variances larger than 3 percent are not widely used in evaluations in previous studies.

\subsubsection{Fast Learning Designs}
\md{We further conduct ablations for exploring the extra loss designs in our fast learning framework. As shown in Tab. \ref{table:fast_ablation}, both the introduced pull loss in Eq. (\ref{eq:pullloss}) and the Eikonal term in Eq. (\ref{eq:eikonal}) contribute to the stable optimization of our fast learning framework based on multi-resolution hash grids. The designed pull loss helps the initializing of field in the areas far away from the surface and the Eikonal term stables the SDFs learning by regularizing the gradients.  }

\begin{table}[h]
\centering
\setlength{\tabcolsep}{3mm}

\resizebox{0.92\linewidth}{!}{
    \begin{tabular}{c|c|c|c}  %
     \hline

          Designs & W/O $\mathcal{L}_{\mathrm{pull}}$ & W/O $\mathcal{L}_{\mathrm{reg}}$ & Full \\   %
     \hline

       L2CD$\times10^4$&8.016&5.255&\textbf{4.537}\\ %
       P2M$\times10^4$&4.209&2.252&\textbf{1.723}\\
       \hline
   \end{tabular}}
   
   \caption{Loss designs in our Fast Learning framework.}
   
   \label{table:fast_ablation}
\end{table}

\begin{figure}[h]
  \centering
   \includegraphics[width=\linewidth]{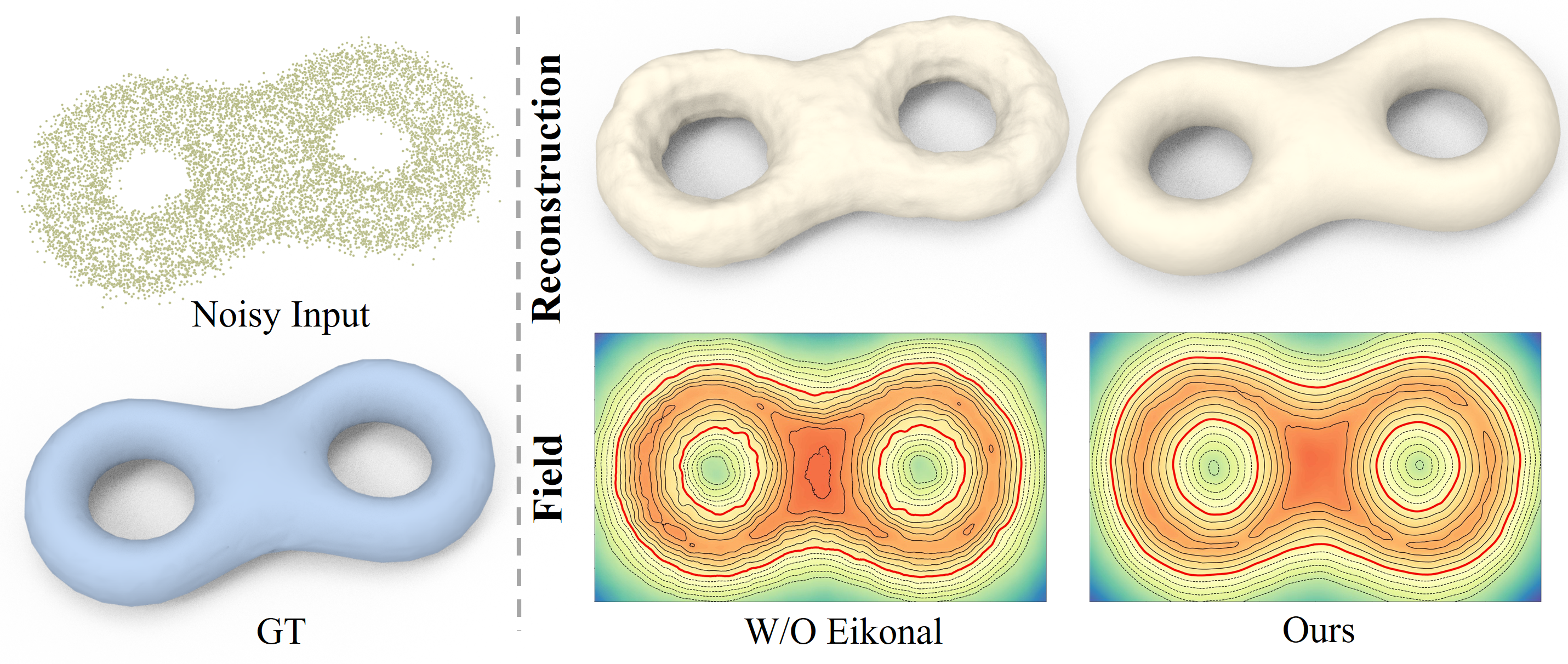}
    \caption{\js{The effect of Eikonal term in our fast learning framework.}}
   \label{fig:eikonal_comp}

\end{figure}
\js{The Eikonal term is necessary for our fast learning framework based on Instant-NGP~\cite{mueller2022instant}, especially when learning SDFs from point clouds with large noise levels. We further provide more illustrations and visualizations on the reconstructions and distance fields of the SDFs learned by our framework with or without Eikonal term. The illustrative visualizations are shown in Fig.~\ref{fig:eikonal_comp}.}

\js{The visual comparisons in Fig.~\ref{fig:eikonal_comp} demonstrate that the Eikonal term plays the key role in stabilizing and regularizing the SDF learning of our fast learning framework. The reconstructions achieve with Eikonal term is significantly smoother and more accurate than the reconstructions achieved without Eikonal term. We further provide the visualization of the distance fields learned with or without Eikonal term in the bottom of Fig.~\ref{fig:eikonal_comp}. The field visualization further demonstrates the effectiveness of Eikonal term in learning SDFs from noisy inputs. The level sets of the distance field learned with Eikonal term are much smoother than the ones of the distance field learned without Eikonal term.}

\subsubsection{Multi-View Reconstruction Designs}
\begin{figure}[h]
  \centering

   \includegraphics[width=0.92\linewidth]{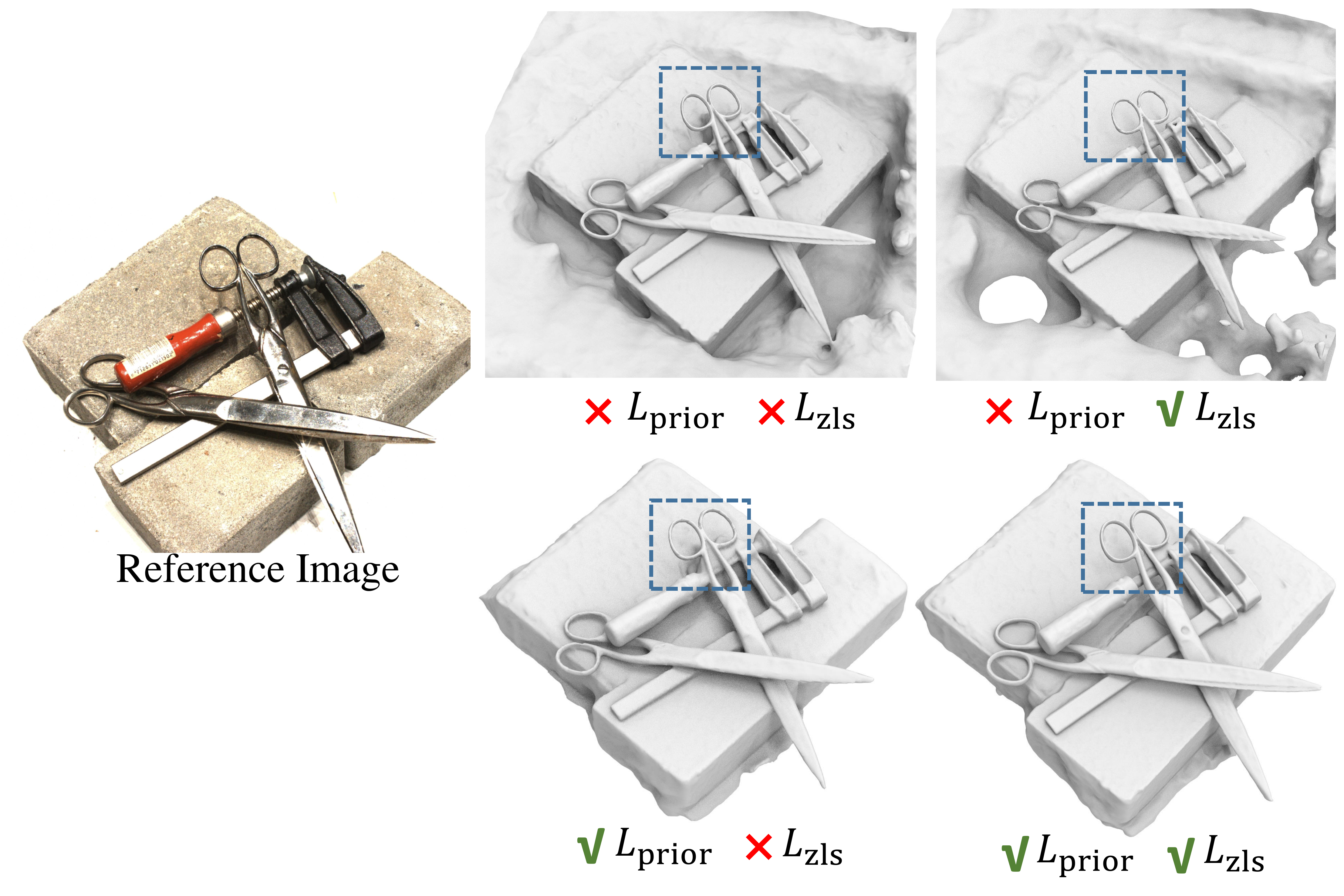}

\caption{\label{fig:mvs_ablation}Ablations on the designed losses in Multi-View Reconstruction task.}
\end{figure}

\md{We then present visualizations of the multi-view reconstruction results under various loss settings to illustrate the effectiveness of our proposed designs. The visualization in Fig. \ref{fig:mvs_ablation} demonstrates that the introduced SDF prior loss $L_{\rm{prior}}$ in Eq. (\ref{eq:prior}) significantly contributes to the reduction of artifacts in empty spaces and slightly improve the reconstruction quality. Conversely, the zero-level set loss $L_{\rm{zls}}$ defined in Equation (\ref{eq:zls}) has limited impact on removing artifacts but substantially helps to capture intricate geometries. The most optimal outcome is achieved through the combined utilization of both loss functions.}

\section{Conclusion}

\label{Conclusion}
We introduce to learn SDFs from noisy point clouds via noise to noise mapping. We explore the feasibility of learning SDFs from multiple noisy point clouds or even one noisy point cloud without the ground truth signed distances, point normals or clean point clouds. Our noise to noise mapping enables the statistical reasoning on point clouds although there is no spatial correspondence among points on different noisy point clouds. Our key insight in noise to noise mapping is to use EMD as the metric in the statistical reasoning. With the capability of the statistical reasoning, we successfully reveal surfaces from noisy point clouds by learning highly accurate SDFs. \md{We also integrate multi-resolution hash encoding implemented in CUDA into our framework, which reduces our training time by a factor of ten, enabling the convergence within one minute. 
Furthermore, we introduce a novel schema to improve multi-view reconstruction by using an SDF prior learned from noisy SfM points. We evaluate our method under synthetic dataset or real scanning dataset for both shapes or scenes. The effectiveness of our method is justified by our state-of-the-art performance in different applications.}

\appendices

\section{Proof}
\label{sec.proof}
We proof Theorem 1 in our submission in the following.

\noindent\textbf{Theorem 1. }\textit{Assume there was a clean point cloud $\bm{G}$ which is corrupted into observations $S=\{\bm{S}_i\}$ by sampling a noise around each point of $\bm{G}$. If we leverage EMD as the distance metric $L$ defined in Eq.~(\ref{eq:4}), and learn a point cloud $\bm{G}'$ by minimizing the EMD between $\bm{G}'$ and each observation in $S$, i.e., $\min_{\bm{G}'}\sum_{\bm{S}_i\in S}L(\bm{G}',\bm{S}_i)$, then $\bm{G}'$ converges to the clean point cloud $\bm{G}$, i.e., $L(\bm{G},\bm{G}')=0$.}
\vspace{-0.1in}

\begin{equation}
\label{eq:4}
\begin{aligned}
L(\bm{G},\bm{G}')=\min_{\phi:\bm{G}\to\bm{G}'}\sum_{\bm{g}\in\bm{G}}||\bm{g}-\phi(\bm{g})\|_2,
\end{aligned}
\end{equation}
\vspace{-0.15in}

\noindent where $\phi$ is a one-to-one mapping.

\noindent\textbf{Proof: }Suppose each corrupted observation $\bm{S}_i$ in the set $S=\{\bm{S}_i|i\in[1,N]\}$ is formed by $m$ points, and $\bm{S}_i=\{n_i^k|k\in[1,m],m\ge 1\}$. With the same assumption, either $\bm{G}$ or $\bm{G}'$ is also formed by $m$ points, $\bm{G}=\{g^k|k\in[1,m],m\ge 1\}$, $\bm{G}'=\{g'^k|k\in[1,m],m\ge 1\}$. Assuming each noise $n_i^k$ is corrupted from the clean $g^k$, we leverage this assumption to justify the correctness of our proof. $L(\bm{G}',S)=\sum_{\bm{S}_i\in S}L(\bm{G}',\bm{S}_i)$.

$(a)$ When $m=1$, this is similar to Noise2Noise~\cite{noise2noise},

\begin{equation}
\begin{split}
\label{eq:Exp}
L(\bm{G}',S)=&\sum_{i=1}^{N}(g'^1-n_i^1)^2. \\
\frac{\partial L(\bm{G'},S)}{\partial \bm{G'}}&=2\sum_{i=1}^{N}(g'^1-n_i^1). \\
\frac{\partial L(\bm{G'},S)}{\partial \bm{G'}}=0 \rightarrow &g'^1 = 1/N\sum_{i=1}^{N}n_i^1.
\end{split}
\end{equation}

Since $S=\{\bm{S}_i\}$ is a set corrupted from the clean point cloud $\bm{G}$, $g^1=1/N\sum_{i=1}^{N}n_i^1$. Furthermore, we also get $g'^1=g^1$.

From Eq.~(\ref{eq:Exp}), we can also get the following conclusion,
\begin{equation}
\label{eq:min}
\mathop{min}\limits_{\bm{G'}}L(\bm{G'},S)\leftrightarrow \bm{G'}=\mathbb{E}(\bm{\phi}(\bm{G'})),
\end{equation}

\noindent where $\bm{\phi}=\{\phi_i|i\in[1,N]\}$ is a set of one-to-one mapping $\phi_i$ which maps $\bm{G'}$ to each corrupted observation $\bm{S}_i$ in $S$.

$(b)$ When $m\ge2$, assuming that we know which noisy point $n_i^k$ on each point cloud $\bm{S}_i$ is corrupted from the clean point $g^k$. We regard the correspondence $c_i$ between $\{n_i^k|i\in[1,N]\}$ and $g^k$ as the ground truth, so that we can verify the correctness of our following proof. Note that we did not use this assumption in the proof process. So, we can represent the correspondence using the following equation,

\begin{equation}
\mathbb{E}(n(k))=1/N\sum_{i=1}^{N}n_i^k=g^k,
\end{equation}

\noindent where $n(k)=\{n_i^k|i\in[1,N]\}$.

As defined before, $\phi_i$ is the one-to-one mapping established in the calculation of EMD between $\bm{G}'$ and $\bm{S}_i$. Therefore, the distance between $\bm{G}'$ and noisy point cloud set $S$ is, $L(\bm{G}',S)=\sum_{k=1}^{m}(\sum_{i=1}^{N}((g'^k-\phi_i(g'^k))^2))$,

There are two cases. One is that the one-to-one mapping $\phi_i$ is exactly the correspondence ground truth $c_i$. The other is that $\phi_i$ is not the correspondence ground truth.

Case $(1)$: When $\phi_i(g'^k)=n_i^k$, $i\in[1,N]$, this is consistent with $(a)$, so the Theorem 1 gets proved.

Case $(2)$: When $\phi_i(g'^k)\neq n_i^k$, assuming $\phi_i(g'^k)=n_i^{a_{k,i}}$, $A_k=\{n_i^{a_{k,i}}|i\in[1,N]\}$, $A_k$ is a set corresponding to $g'^k$. When minimizing $L(\bm{G}',S)=\sum_{k=1}^{m}\sum_{i=1}^{N}(g'^k-\phi_i(g'^k))^2$, according to Eq.~(\ref{eq:min}), $g'^k=\mathbb{E}(\phi_i(g'^k))$, so $Var(A_k)=1/N\sum_{i=1}^{N}((g'^k-\mathbb{E}(\phi_i(g'^k)))^2)$.
When $m=2$, $\mathop{min}\limits_{\bm{G'}}L(\bm{G'},S)=\min(Var(A_1)+Var(A_2))$. We assume $A_1=n_s^1+n_{cs}^2$ to simply the following proof, where $s$ is a subset of set $[1,N]$, ${cs}$ is the complement of set $s$, so $A_2=n_s^2+n_{cs}^1$. Assuming $\mathbb{E}(A_1)=g^1+\Delta$, $\Delta$ is the point offset of $g^1$, because of $\mathbb{E}(A_1)+E(A_2)=g^1+g^2$, so $\mathbb{E}(A_2)=g^2-\Delta$,

\begin{equation*}
\begin{split}
L(\bm{G'},S)\\
=&(Var(A_1)+Var(A_2))\\
   =& \mathbb{E}(A_1-(g^1+\Delta))^2+\mathbb{E}(A_2-(g^2-\Delta))^2 \\
    =& 1/N(\sum_{i=1}^{N}(n_i^{a_{1,i}})^2+\sum_{i=1}^{N}(n_i^{a_{2,i}})^2+N(g^1+\Delta)^2\\
    &+N(g^2-\Delta)^2-2\sum_{i=1}^{N}n_i^{a_{1,i}}(g^1+\Delta)\\
    &-2\sum_{i=1}^{N}n_i^{a_{2,i}}(g^2-\Delta)) \\
    =& \mathbb{E}((n(1))^2)+\mathbb{E}((n(2))^2)+\mathbb{E}^2(n(1))+\\
    &\mathbb{E}^2(n(2))+2\Delta^2+2g^1\Delta-2g^2\Delta-\\
    &2/N(g^1\sum_{i=1}^{N}n_i^{a_{1,i}}+g^2\sum_{i=1}^{N}n_i^{a_{2,i}}+\\
    &\Delta\sum_{i=1}^{N}n_i^{a_{1,i}}-\Delta\sum_{i=1}^{N}n_i^{a_{2,i}}) \\
    =&\mathbb{E}((n(1))^2)+\mathbb{E}((n(2))^2)+\mathbb{E}^2(n(1))+\\
    &\mathbb{E}^2(n(2))+2/N(\Delta(n_s^1+n_{cs}^1)-\Delta(n_s^2+n_{cs}^2)\\
    &-\Delta(n_s1+n_{cs}^2)+\Delta(n_{cs}^1+n_s^2)- \\
    &g^1\sum_{i=1}^{N}n_i^{a_{1,i}}-g^2\sum_{i=1}^{N}n_i^{a_{2,i}})\\
    =& \mathbb{E}((n(1))^2)+\mathbb{E}((n(2))^2)+\mathbb{E}^2(n(1))+\\
    &\mathbb{E}^2(n(2))+2\Delta^2+{2\Delta}/N(2n_{cs}^1-\\
    &2n_{cs}^2)-2/N(g^1N(g^1+\Delta)+g^2N(g^2-\Delta)) \\
    =& \mathbb{E}((n(1))^2)+\mathbb{E}((n(2))^2)-\mathbb{E}^2(n(1))-\\
    &\mathbb{E}^2(n(2))+2\Delta^2+\\
    &{2\Delta}/N(2n_{cs}^1-2n_{cs}^2-n_{cs}^1-n_s^1+n_{cs}^2+n_s^2) \\
    =& \mathbb{E}((n(1))^2)+\mathbb{E}((n(2))^2)-\mathbb{E}^2(n(1))-\\
    &E^2(n(2))+2\Delta(g^2-g^1)-2\Delta^2 \\
    =& Var(n(1))+Var(n(2))+2\Delta(g^2-g^1)-2\Delta^2 \\
\end{split}
\end{equation*}

Because the first two terms of the formula are constants, the entire formula becomes a quadratic formula, so when $\Delta=0$ or $\Delta=g^2-g^1$, the value of $L(\bm{G'},S)$ is minimized. $\Delta=0$ is consistent with Case $(1)$. $\Delta=g^2-g^1$, $\phi_i(g^{1})=n_i^2$, $\phi_i(g^{2})=n_i^1$, this is also the same correspondence as the ground truth, so Theorem 1 gets proved. When $m\textgreater2$. We can extend the proof from the two sets $A_1$ and $A_2$ to multiple sets $A_1,A_2,\cdots,A_m$, and the proof process is similar to the above.

\section{More Experiments and Explorations}

\begin{figure}[tb]
  \centering
   \includegraphics[width=\linewidth]{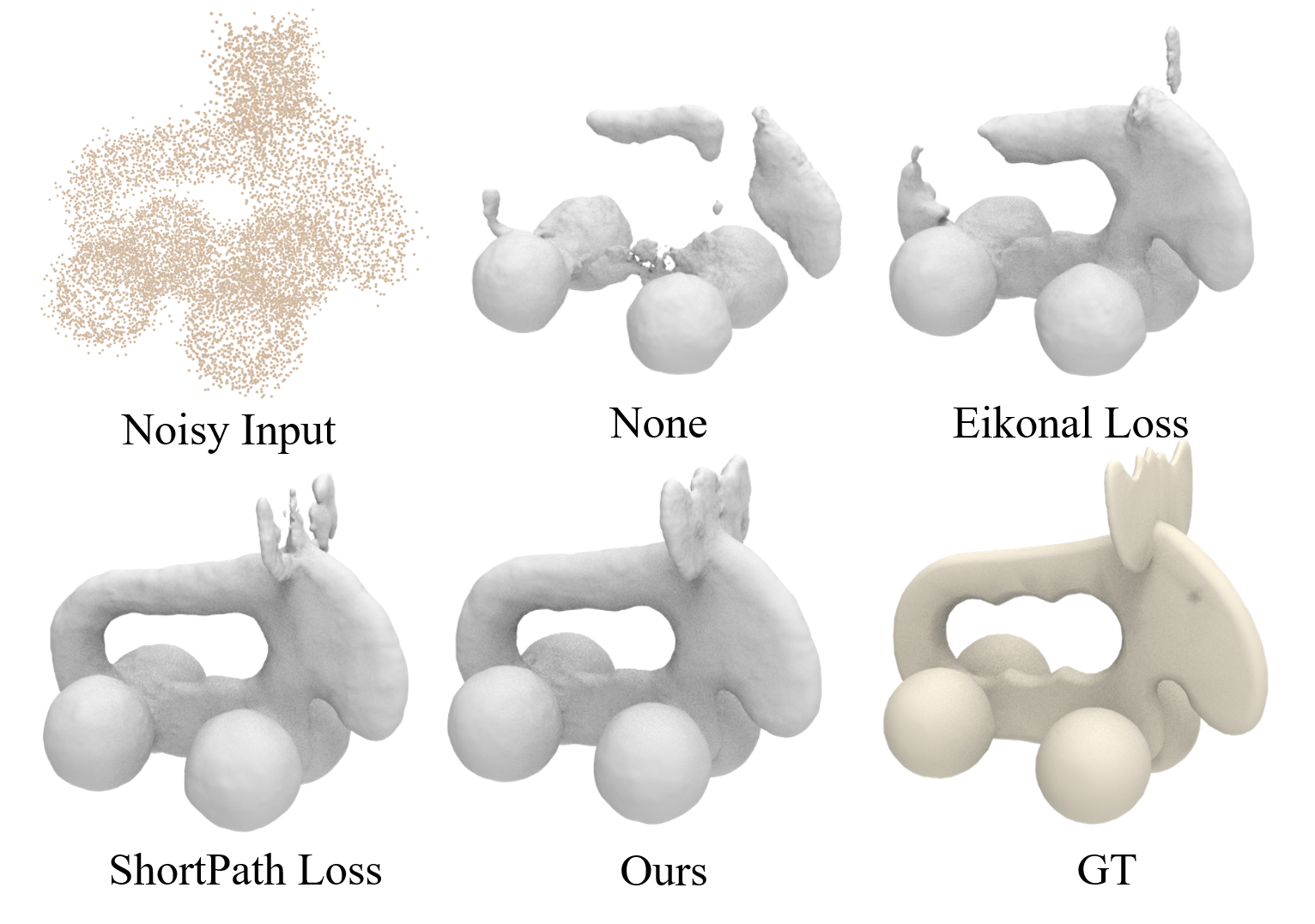}
    \caption{\js{The effect of different constraints on geometric consistency.}}
   \label{fig:ablation_gc}

\end{figure}

\begin{table}[!]
\setlength\tabcolsep{1.5mm}

\centering
\resizebox{0.9\linewidth}{!}{
    \begin{tabular}{c|c|c|c|c}  %
     \toprule
          &None&Eikonal Loss &ShortPath Loss& Ours\\
     \midrule
       CD&80.917&8.616&5.483&\textbf{4.277}\\ %
       P2M&- & 5.109 & 2.435 & \textbf{1.847}\\
       \bottomrule
   \end{tabular}}
   
   \caption{\js{Surface reconstruction comparisons of regularizing with Eikonal loss, ShortPath loss and the proposed geometry consistency loss $R$ under PUNet dataset \cite{DBLP:conf/cvpr/YuLFCH18}.}}
   \label{table:ablation_gcloss}
\end{table}

\subsection{Alternatives in Geometric Consistency}
\js{We conduct ablation studies for exploring the superiority of our proposed Geometric Consistency Constraint compared with the alternative implementations of a) Eikonal loss \cite{gropp2020implicit}, or b) encouraging short paths for moving queries in Tab.~\ref{table:ablation_gcloss}.}

\js{Specifically, we employ an Eikonal term with a lower coefficient for regularizing the learning of SDF from noisy point clouds and show the results as ``Eikonal Loss''. To encourage short paths for moving queries, we simply add a constraint to minimize the absolute value of the predicted signed distances at each query location and report the results as “ShortPath Loss”. ``None'' indicates that no additional constraints on the geometric consistency are used during training. As shown in Tab.~\ref{table:ablation_gcloss} and Fig.~\ref{fig:ablation_gc}, our designed geometric consistency loss achieves the best performance over all the three constraints. }

\js{The results also demonstrate the effectiveness of the alternative constraints in improving the reconstruction qualities. Specifically, the results are degenerated without proper constraints on geometric consistency. By introducing the Eikonal term for regularizing SDF, the performances are much more convincing, where the CD reduces by an order of magnitude. However, the Eikonal loss which implicitly constrains on the learning of SDFs by constraining gradients, does not always bring stable enough optimizations for the difficult situations where only noisy point clouds are available. A more explicit and powerful constraint on the SDFs is required for further improvement. }

\begin{figure*}[t]
  \centering
   \includegraphics[width=\linewidth]{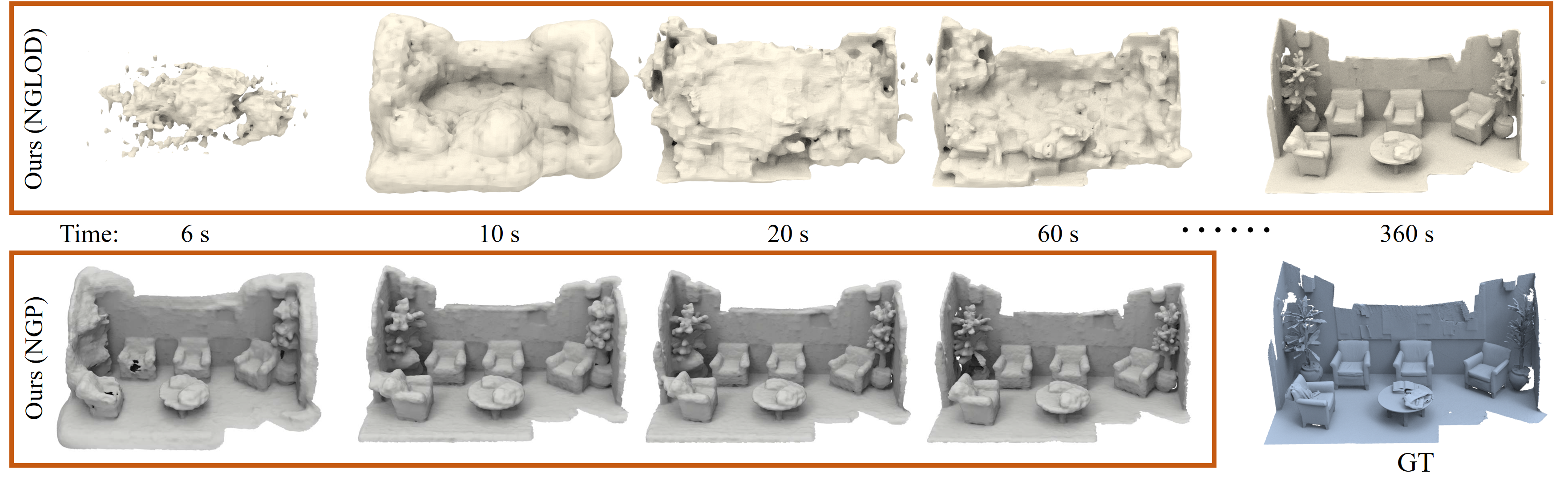}
    \caption{\js{The comparisons on the efficiency of different fast learning frameworks.}}
   \label{fig:nglod_n2n}

\end{figure*}

\js{The ShortPath loss is a more explicit and direct term for regularizing SDF and therefore leads to more robust surface reconstruction from noisy point clouds by encouraging the moving distances to be minimal. However, the ShortPath loss constrains on each point equally, which is not aware of the far/near space and does not provide the exact target for how far the moving path should be. Moreover, an improper weight of ShortPath loss may prevent some queries in the far space to move correctly to the surface. The best performance is achieved with our proposed geometric consistency constraint, which explicitly provides an exact and accurate optimization target for each query point and leads to stable and robust surface reconstructions. }

\subsection{Alternatives in Fast Learning Framework}

\js{We explore the fast learning framework based on Instant-NGP \cite{mueller2022instant}, which has been proven to be an extremely efficient representation for implicit functions. To demonstrate the effectiveness of choosing Instant-NGP for fast learning, we further implement another efficient implicit representation architecture NGLOD \cite{takikawa2021neural} on our noise-to-noise learning framework by adopting the octree-based feature volume used in NGLOD for representing SDF.  We show the visualizations on the optimization process of our method with NGLOD and Instant-NGP as the fast learning frameworks in Fig.~\ref{fig:nglod_n2n}.}

\js{The visualizations in Fig.~\ref{fig:nglod_n2n} demonstrate that adapting Instant-NGP achieves significantly faster convergence compared to the NGLOD-based implementation in our framework. But we further justify that the NGLOD representation is also effective in improving the training efficiency of our method, which reduces the training time from 15 minutes in the original version of our method to 6 minutes on the complex 3DScene~\cite{DBLP:journals/tog/ZhouK13} dataset. While our NGP-based fast learning framework only needs about 60 seconds.}

\subsection{Integrating Fast learning Framework with Other Methods}

\begin{figure*}[tb]
  \centering
   \includegraphics[width=0.9\linewidth]{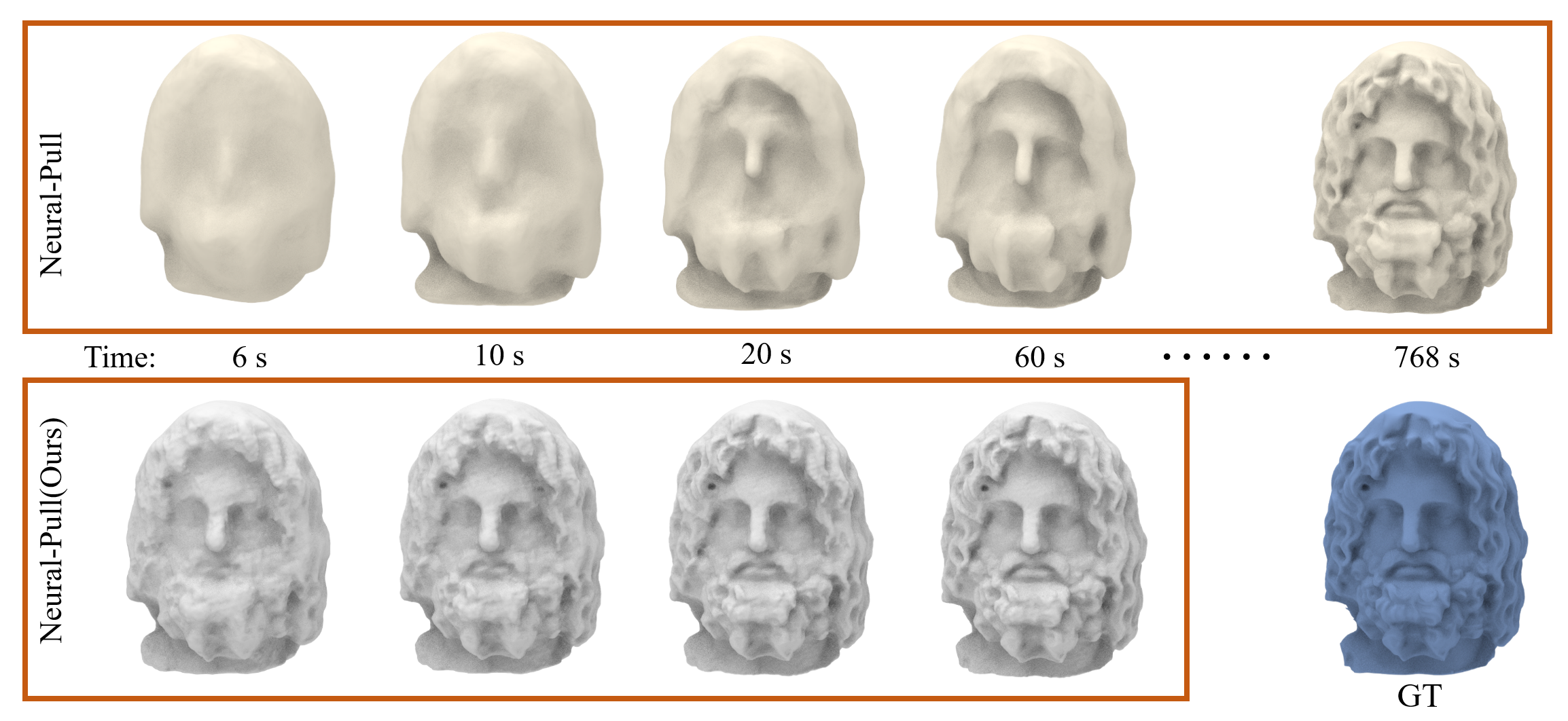}
    \caption{\js{Improving the learning efficiency of Neural-Pull by integrating the fast learning framework.}}
   \label{fig:np_fast}

\end{figure*}

\js{We admit that our fast learning framework based on Instant-NGP is not limited to the noise-to-noise learning schema and also has the potential to improve the efficiency of other methods on reconstructing surfaces from point clouds. We integrate the fast learning framework to Neural-Pull \cite{ma2020neuralpull} and conduct qualitative comparisons at different time of the process of reconstructing surfaces from the ``Serapis'' model. The visual comparisons are shown in Fig.~\ref{fig:np_fast}.}

\js{Since Neural-Pull can not handle noisy point clouds, we densely sample 100k clean points from the ``Serapis'' model and train the original Neural-Pull and our implemented fast learning version of Neural-Pull to reconstruct surfaces from the clean point clouds. As shown in Fig.~\ref{fig:np_fast}, the fast learning framework works well on other methods, which significantly accelerates the convergence of Neural-Pull by more than 10 times.}

\subsection{Efficiency Comparisons with SOTA Fast Learning Researches}
\begin{figure*}[tb]
  \centering
   \includegraphics[width=\linewidth]{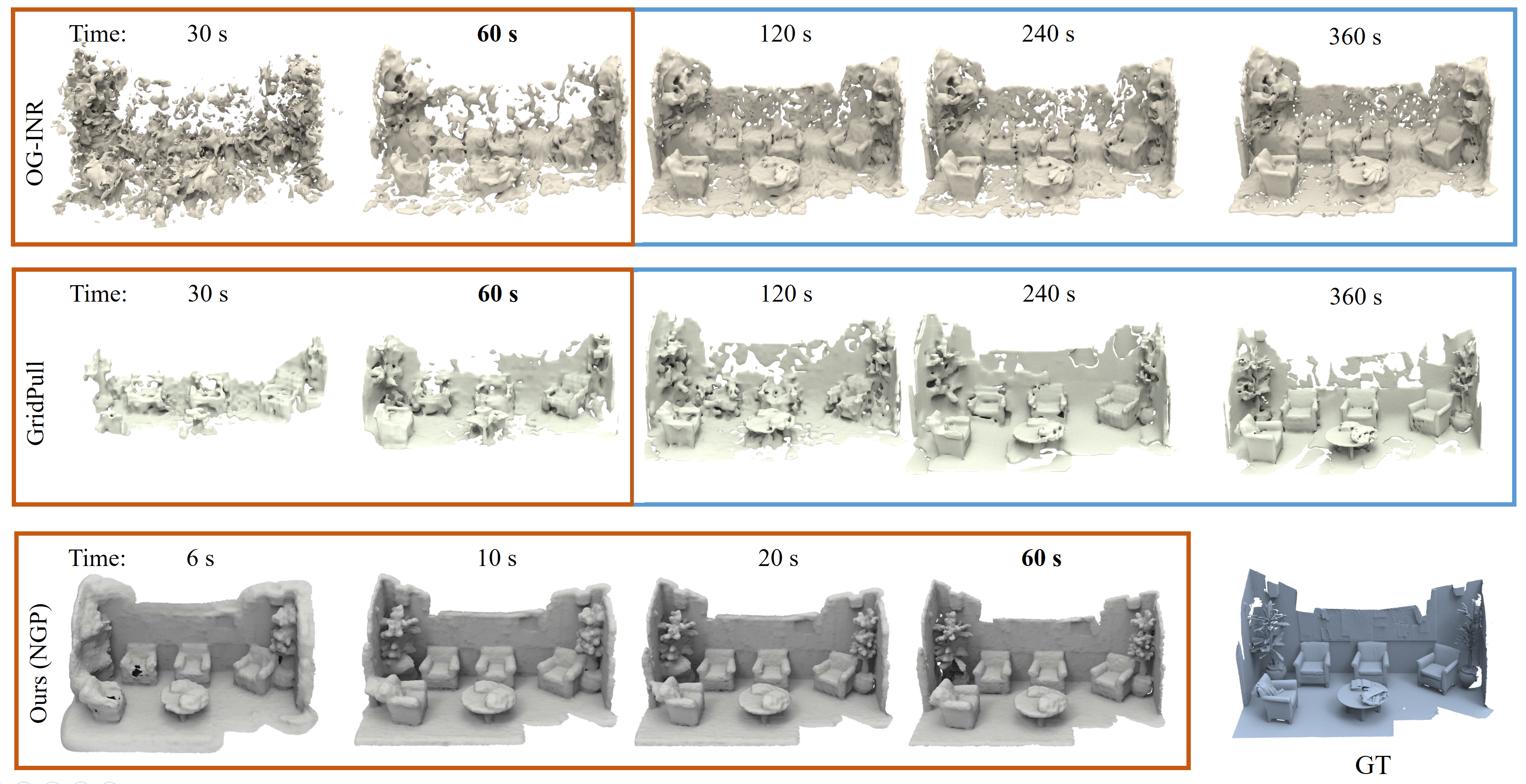}
    \caption{\js{The efficiency comparisons with SOTA fast learning researches OG-INR and GridPull.}}
\label{fig:fast_comp}
\end{figure*}
\js{We further conduct evaluations to compare our method with OG-INR \cite{koneputugodage2023octree} and GridPull \cite{chen2023gridpull} to demonstrate the efficiency and effectiveness of our fast learning framework. We visually compare the optimization process in Fig.~\ref{fig:fast_comp}. The results of OG-INR and GridPull are reproduced with their official codes and settings under a single 3090 GPU. Note that OG-INR additionally requires a long time for building octree, which we exclude the time for intuitive visualizations on the optimization process. The time to build octree for OG-INR differs with different point cloud scales. For the simple shapes in ShapeNet, it may only take 40 seconds as the authors reported in Tab.3 of their paper. For the complex scene we used here, the time increases greatly to about 8 minutes. }

\js{As shown in Fig.~\ref{fig:fast_comp}, our method requires significantly less time for convergence when compared with OG-INR and GridPull. Note that the time for OG-INR, GridPull and our method at each column in the figure is not aligned. This is for a clearer visualization for the optimization process of each method since our method requires much less time. For the complex scene, OG-INR and GridPull require about 4-6 minutes to converge, while our fast learning framework only takes 20-60 seconds to learn an accurate SDF for surface reconstruction. Moreover, the two fast-learning baselines struggle to produce clean and complete surfaces, whereas our method can accurately reconstruct high-fidelity scenes with greater efficiency.}

\subsection{Effect of Our SDF Prior for Guiding NeuS}
\begin{figure}[tb]
  \centering
   \includegraphics[width=\linewidth]{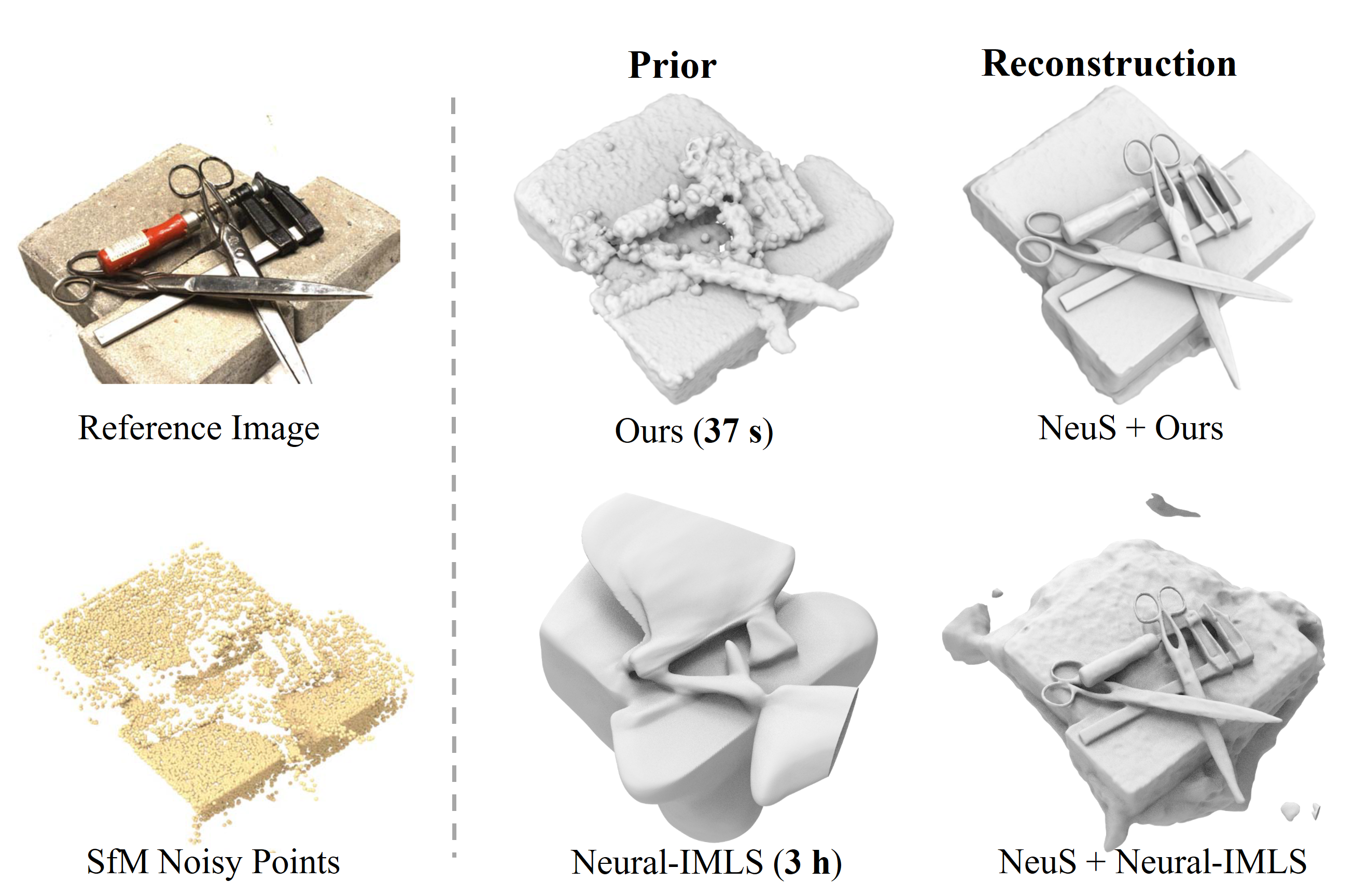}
    \caption{\js{The visualizations of SDF priors and multi-view reconstructions of NeuS guided by the priors produced by our method and Neural-IMLS.}}
   \label{fig:nimls_comp}

\end{figure}
\js{Our method which learns robust and stable signed distance functions (SDFs) from noisy and corrupted point clouds, is a key factor in enhancing the convergence and performance of multi-view reconstruction methods (e.g., NeuS \cite{neuslingjie}). The explicit point clouds obtained with Structure from Motion (SfM) methods (e.g. COLMAP \cite{schoenberger2016sfm}) often suffer from noises with unknown distributions, where the previous methods struggle in learning robust SDFs from them. To demonstrate the superiority of our method to serve as the prior for multi-view reconstruction, we conduct experiments to compare the multi-view surface reconstruction qualities achieved with the guidance of SDF prior learned by our method and the one learned by Neural-IMLS \cite{wang2021neural}.  The visualizations of SDF priors and multi-view reconstructions of NeuS guided by the priors produced by our method and Neural-IMLS are shown in Fig.~\ref{fig:nimls_comp}.}

\js{As shown in Fig.~\ref{fig:nimls_comp}, our method significantly outperforms Neural-IMLS in learning accurate and stable SDF priors from the noisy point clouds obtained by SfM, which further contributes to the clean and accurate multi-view reconstructions by guiding NeuS’s learning. The reason is that Neural-IMLS struggles in handling the noisy and corrupted point clouds obtained by SfM. }

\js{Moreover, our method is significantly more efficient than Neural-IMLS. We adopt the official code and settings of Neural-IMLS for learning SDFs, where the training for one shape takes about 10 hours in total using a single 3090 GPU. We also found that Neural-IMLS suffers from overfitting when learning SDF from noisy and corrupted point clouds. To explore the upper limit of Neural-IMLS prior for guiding NeuS’s learning, we adopt the best SDF learned by Neural-IMLS which is achieved at the third hour as the prior for implementing prior-based NeuS with the guidance of Neural-IMLS. The results have been reported in Fig.~\ref{fig:nimls_comp}.}

\subsection{Ablations on Guiding NeuS with SDF Prior}
\begin{figure}[tb]
  \centering
   \includegraphics[width=\linewidth]{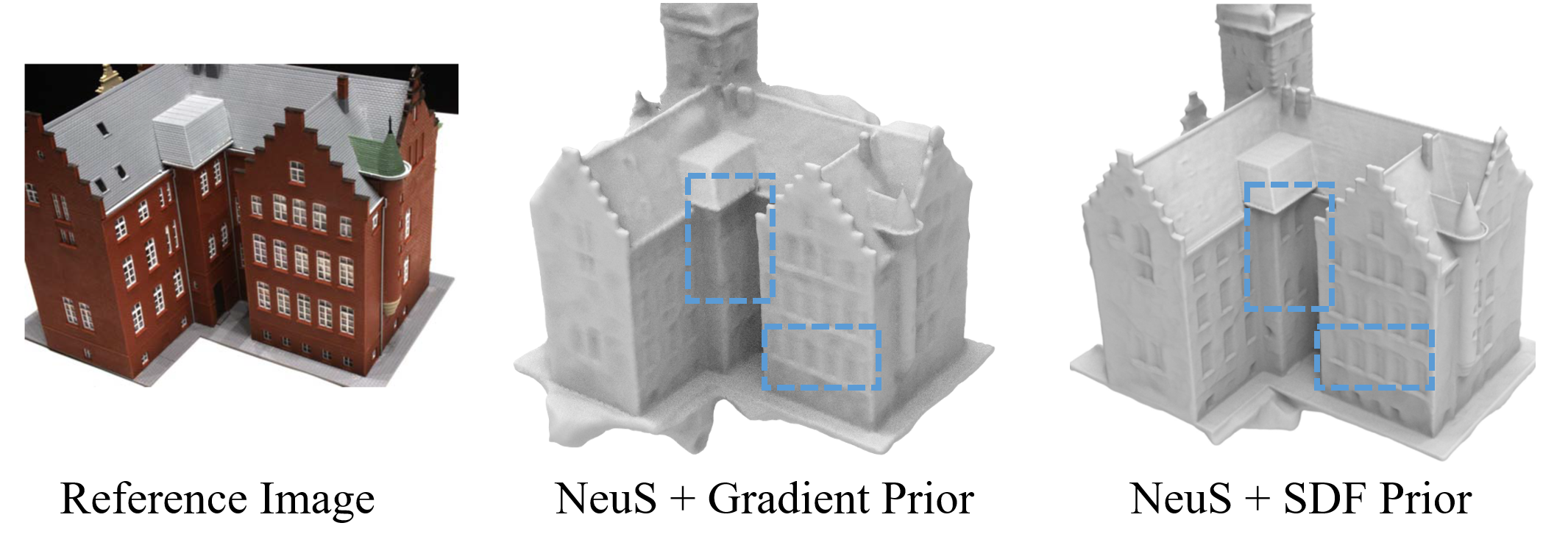}
    \caption{\js{The visualizations of SDF priors and multi-view reconstructions of NeuS guided by the priors produced by our method and Neural-IMLS.}}
   \label{fig:gradient_prior}

\end{figure}
\js{We implement a new framework for guiding the learning of NeuS \cite{neuslingjie} by utilizing the gradients of our learned SDF prior to constrain the SDF gradient of NeuS. This is achieved by encouraging the consistency between the query gradients of our learned SDF prior and the query gradients of the learning SDF of NeuS. We show the visual comparison of two implementations in Fig.~\ref{fig:gradient_prior}.}

\js{As shown, the simple implementation which directly takes signed distances as the prior for guiding NeuS learning achieves better performance in producing detailed geometries. Leveraging gradients of our learned SDF as a guidance also works in removing artifacts and reconstructing cleaner and smoother surfaces compared to the original NeuS. However, the indirect supervision by constraining gradients is not as effective as constraining at the distance values, and thus struggles to preserve high-fidelity details.}

\subsection{Comparison with DiGS}
\begin{figure}[tb]
  \centering
   \includegraphics[width=\linewidth]{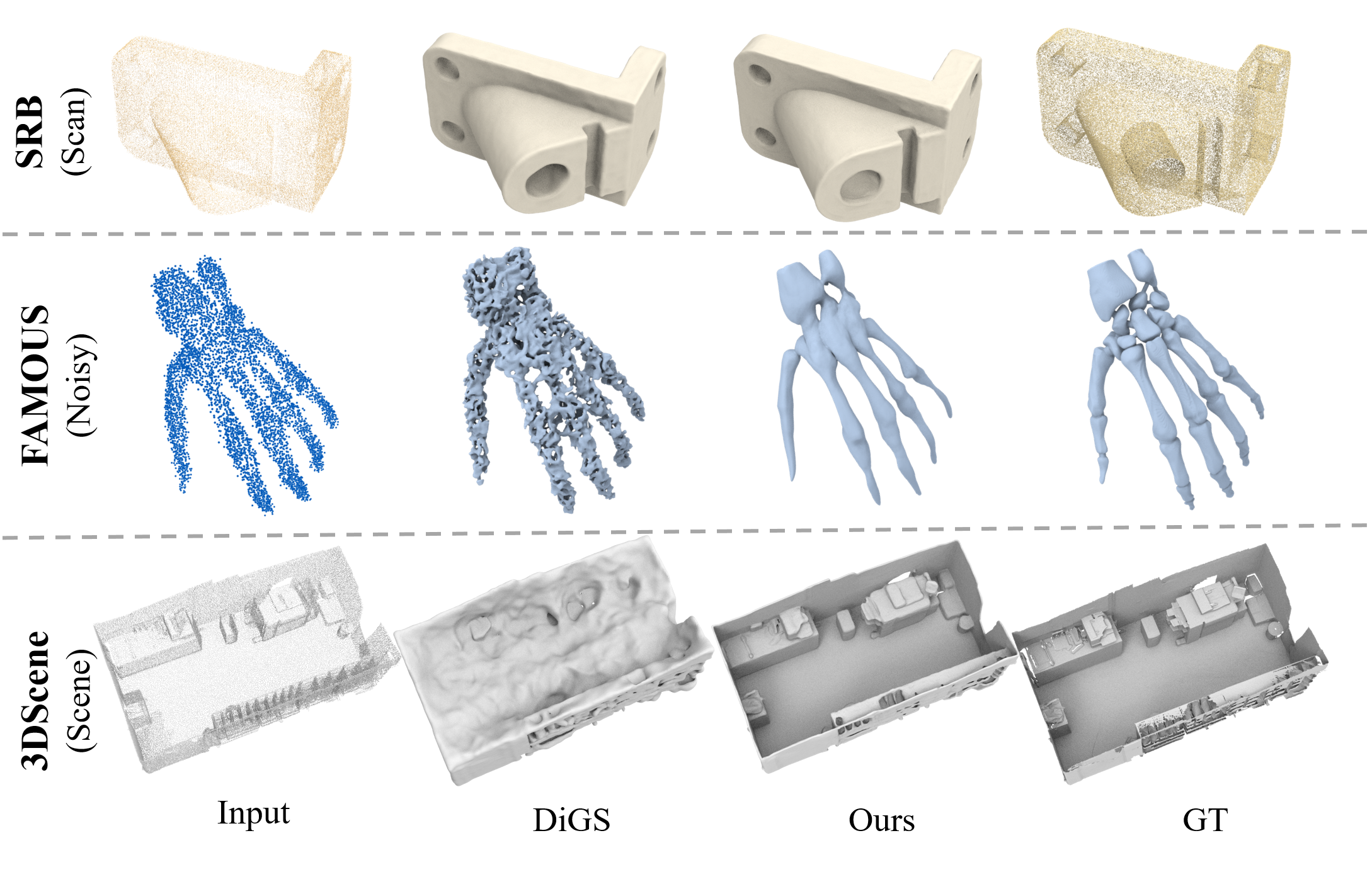}
    \caption{\js{Comprehensive comparisons with SOTA overfitting based method DiGS under scanned point clouds, noisy point clouds and scene-level point clouds.}}
   \label{fig:digs_comp}

\end{figure}
\js{For a comprehensive comparison with SOTA overfitting based method DiGS \cite{ben2021digs}, we further provide the visual comparisons with DiGS under SRB \cite{berger2013benchmark}, FAMOUS and 3DScene~\cite{DBLP:journals/tog/ZhouK13} dataset to evaluate the performance under scanned point clouds, noisy point clouds and scene-level point clouds. All the results are achieved with the official code and settings provided by DiGS. For SRB and FAMOUS dataset, we leverage the object reconstruction setting of DiGS. For 3DScene dataset, we leverage the provided scene reconstruction setting. The results are shown in Fig.~\ref{fig:digs_comp}.}

\js{DiGS achieves convincing performance in the SRB dataset, which demonstrates its superiority in object-level surface reconstruction. However, DiGS fails to reconstruct clean surfaces for noisy point clouds and also struggles in handling the complex scenes in the 3DScene dataset. On the contrary, our method robustly reconstructs clean and high-fidelity surfaces from real-scanned point clouds, noisy point clouds and complex scene point clouds.  }

\subsection{Comparison with POCO}
\begin{figure}[tb]
  \centering
   \includegraphics[width=\linewidth]{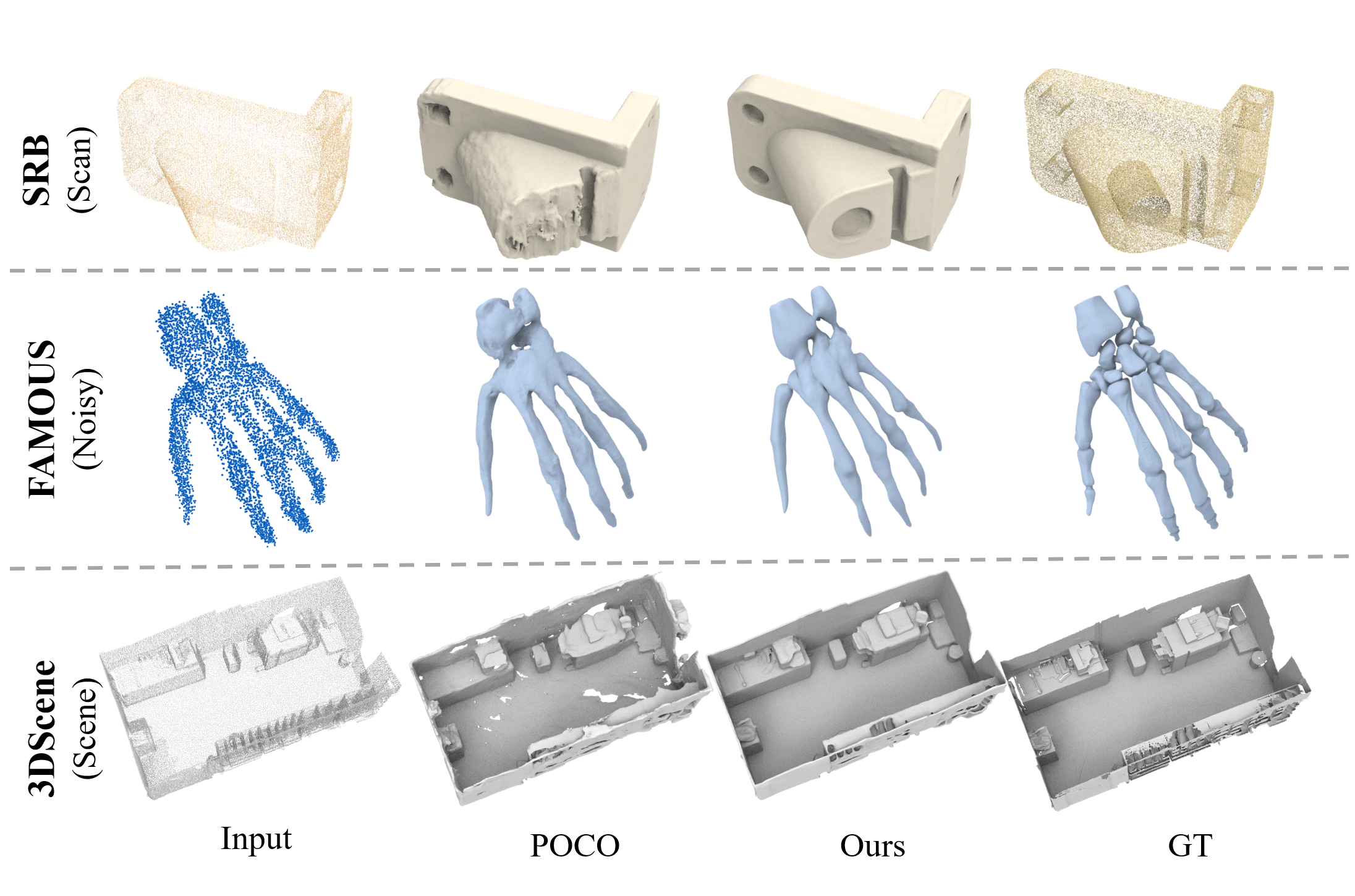}
    \caption{\js{Comprehensive comparisons with data-driven supervised method POCO under scanned point clouds, noisy point clouds and scene-level point clouds.}}
   \label{fig:poco_comp}

\end{figure}

\js{For a comprehensive comparison with SOTA data-driven supervised method POCO~\cite{pococvpr2022}, we further provide the quantitative comparisons with POCO under SRB \cite{berger2013benchmark}, FAMOUS and 3DScene~\cite{DBLP:journals/tog/ZhouK13} dataset to evaluate the performance under scanned point clouds, noisy point clouds and scene-level point clouds. All the results are achieved with the official code, settings and pretrained models provided by POCO. For SRB and FAMOUS dataset, we leverage the object reconstruction setting of POCO with the learned model pretrained under ABC dataset. For 3DScene dataset, we leverage the provided scene reconstruction setting with the learned model pretrained under SyntheticRooms~\cite{Peng2020ECCV} dataset. The results are shown in Fig.~\ref{fig:poco_comp}.}

\js{After time-consuming training under large-scale datasets with ground truth occupancy values as supervisions, POCO demonstrates good ability in handling noisy point clouds and scene-level point clouds. However, POCO struggles to produce high-fidelity reconstructions. As shown in the second and third row of Fig.~\ref{fig:poco_comp}, the reconstructions of POCO can not represent the detailed geometries (e.g. the joints of the hand) and also fail in producing complete and accurate geometries for complex scenes (e.g. the holes in the room). Moreover, when leveraging POCO to reconstruct surfaces from the out-of-distribution data of real-scanned point clouds in SRB dataset, the results degenerate dramatically. The reason is that the data-driven based methods which are trained under specific datasets, may struggle in generalizing to point clouds in other domains. On the contrary, our method which even does not require any ground truth supervisions, robustly reconstructs clean and high-fidelity surfaces from real-scanned point clouds, noisy point clouds and complex scene point clouds.  }

\subsection{More Comparisons with SOTA Data-driven Supervised Methods}
\begin{figure}[tb]
  \centering
   \includegraphics[width=\linewidth]{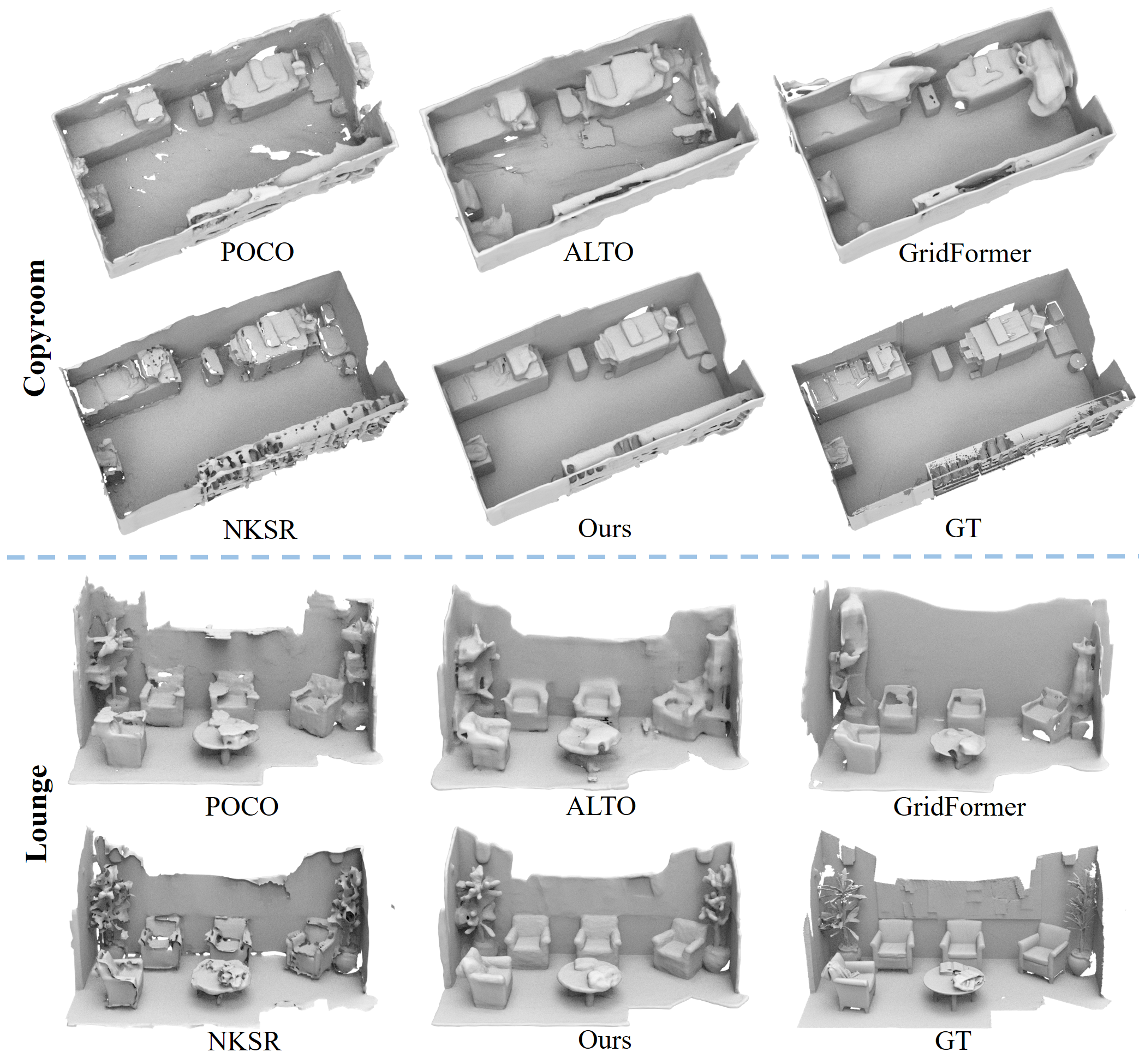}
    \caption{\js{Visual comparisons with SOTA data-driven supervised methods under 3DScene dataset.}}
   \label{fig:3dscene_comp_more}

\end{figure}

\js{To demonstrate the superiority of our method in reconstructing surfaces for scene-level point clouds, we further conduct experiments to make comparisons with SOTA supervised methods POCO~\cite{pococvpr2022}, ALTO~\cite{wang2023alto}, GridFormer~\cite{li2024gridformer} and NKSR~\cite{huang2023neural}. The quantitative comparison results are shown in the main paper. We adopt their official code and the provided pretrained models under scene-level datasets to reproduce their results. Note that NKSR requires point normals as additional inputs for reconstructing surfaces, where we follow their instructions to estimate normals before inferencing. }

\js{We further visually compare our method with these SOTA data-driven methods with 3D supervisions in Fig.~\ref{fig:3dscene_comp_more}. Both the quantitative and qualitative comparisons demonstrate that our method, which does not require any ground truth supervision, outperforms previous supervised methods trained on large-scale datasets. NKSR achieves comparable performance with our method. However, NKSR additionally requires ground truth normals during training, which are not used in our method and other supervised baselines.}

\bibliographystyle{IEEEtran}
\bibliography{egbib}

\end{document}